%% file: latex/0_acl_latex.tex
\pdfoutput=1

\documentclass[11pt]{article}

\usepackage[preprint]{acl}

\usepackage{times}
\usepackage{latexsym}

\usepackage[T1]{fontenc}

\usepackage[utf8]{inputenc}

\usepackage{microtype}

\usepackage{inconsolata}

\usepackage{graphicx}

\usepackage{tcolorbox}
\tcbuselibrary{breakable}

\usepackage{placeins}

\usepackage{enumitem}
\usepackage{calc}
\usepackage{amsmath}

\usepackage{tabularx} 

\usepackage{booktabs}

\usepackage{enumitem}

\title{Narrative-UFET: Narrative Generation for Ultra-Fine Entity Typing}

\author{
Mreedul Gupta \quad Advait Deshmukh \quad Ashwin Umadi \quad Matt Pauk \quad Maria Leonor Pacheco \\
University of Colorado Boulder \\
\texttt{\{mreedul.gupta, advait.deshmukh, ashwin.umadi, matt.pauk, maria.pacheco\}@colorado.edu}
}

\begin{document}
\maketitle
\begin{abstract}
Ultra-fine entity typing (UFET) assigns highly specific types to entity mentions, but current approaches struggle with types in the long tail. We hypothesize that a key limitation is the reliance on sentence-level context, since disambiguating evidence is often spread across multiple sentences. Testing this has been difficult because all existing UFET resources are sentence-level. We present Narrative-UFET, a controlled extension of UFET in which each entity mention is paired with an automatically generated short, coherent narrative. Synthesizing narratives lets us isolate the effect of specific discourse properties. We experiment with two paired variants: one in which the entity's type is held constant across the narrative (\textit{Maintain}) and one in which it shifts (\textit{Change}). We show that narrative context yields consistent improvements on long-tail types over sentence-level baselines, with the \textit{Change} variant providing the stronger signal. A comparison against naturally occurring contexts shows that synthetic narratives yield stronger gains, indicating that controlled discourse construction can surface signals that real text leaves implicit. Substantial room for improvement remains, suggesting open directions in both discourse modeling and narrative construction. 

\end{abstract}

\input{latex/1_intro}

\input{latex/2_related}

\input{latex/3_dataset}
\input{latex/4_experiments}

\input{latex/5_conclusion}

\input{latex/6_limitations}

\input{latex/7_ethical-considerations}


\appendix

\section{Experimental Details for Narrative Generation}
In this section all experimental outputs were generated using models accessed through Ollama, using the default inference settings provided by Ollama for each model. All models were run in quantized form, with GPT-OSS-20B using MXFP4 quantized, while Llama3.3-70B, Gemma3-27B, Qwen3-8B, Qwen3-14B, Qwen3-32B, and Mistral-7B using Q4\_K\_M quantization. Each model kept its default temperature setting. We list all prompts used in narrative generation, including those for model testing, narrative evaluation, prompt testing, and final narrative generation. Prompt variants are shown in a consolidated form, as most variants only differ minimally. The placement of prompt variants reflects where they are input to the model. However, in our experiments each variant was run individually.

\subsection{Model Testing}
\label{app:model_testing}
\subsubsection{Prompt}
\label{app:model_testing_prompt}
To select the best model for our final narrative generations the following prompt was used.

\begin{tcolorbox}[
    colback=gray!10,
    colframe=black,
    width=\columnwidth,
    breakable
]
\ttfamily
You are writing a short, coherent story (<15 sentences) about the entity: "\{entity\}". \\

Rules: \\
- The sentence below MUST appear word-for-word, exactly as written in the narrative. \\
- You cannot paraphrase, reword, or change punctuation in that sentence. \\
- Wrap every mention of the entity "\{entity\}" in <e></e> tags. \\
- The story should naturally include the sentence and expand around it. \\

MANDATORY SENTENCE (must appear exactly as shown): \\
"\{entity\_sentence\}"
\end{tcolorbox}

\subsubsection{Generation Example}
\label{app:narrative_example}

Given the entity \textit{She} and the given sentence \textit{She said the Broad - Burkle offer takes on a ‘shocking’ amount of debt and the Chandlers, despite their bid, do not really run the business”}, a generated narrative would be:

\begin{tcolorbox}[
    colback=gray!10,
    colframe=black,
    width=\columnwidth,
    breakable
]
\ttfamily
\textit{“The boardroom hummed with tension as \textbf{she} adjusted her glasses, her voice cutting through the murmurs. \textbf{She} said the Broad - Burkle offer takes on a “shocking” amount of debt and the Chandlers, despite their bid, do not really want to run the business. A junior analyst choked on her coffee, while the CEO’s jaw tightened. \textbf{She} leaned forward, charts spread across the table, each number a dagger to the room’s fragile optimism. “The Chandlers are bidding to inflate the price,” \textbf{she} added, “not to steward the company.” The chairperson sighed, rubbing his temples. “And your recommendation?” \textbf{she} paused, then said, “We walk.” The room erupted in chaos. Later, alone in the elevator, \textbf{she} replayed the meeting, wondering if truth could ever feel less like a weapon. By dawn, the Chandlers had withdrawn. The Broad-Burkle deal collapsed. \textbf{She} never returned to that boardroom, but the quote lingered, etched into the company’s lore.”}
\end{tcolorbox}

\subsection{Prompt for TinyStories Evaluation Framework}
\label{app:tiny_stories_eval_framework}

The following prompt is a modified version of the TinyStories evaluation framework evaluated on 'Grammar,' 'Creativity,' 'Consistency,' and 'Plot.' Due to model constraints GPT-4 was swapped with GPT-OSS-20B.
\begin{tcolorbox}[
    colback=gray!10,
    colframe=black,
    width=\columnwidth,
    breakable
]
\ttfamily
Your task is to evaluate how well the narrative was written. \\ 
- The narrative is: \{narrative\} \\
- The story should contain the following sentence: \{entity\_sentence\}. \\
\\
Grade the narrative in terms of grammar, creativity, consistency, and whether the plot makes sense. Use the following grading format (exactly): \\
- Grammar: x/10, Creativity: x/10, Consistency: x/10, Plot: x/10
\end{tcolorbox}

\subsection{Discourse Coherence Implementation Details}
\label{app:discourse_coherence}
Discourse Coherence is measured using a bert-based-uncased model which produces a sentence-level embedding vector, mapping each sentence. Cosine similarity is then ran on the vectors where a value of 1 indicates high relatedness and a value of 0 indicating low relatedness. 

\subsection{Number of Characters}
\label{app:num_of_characters}
To study the effect of character constraints on narrative quality, we experiment with prompts that have a specified number of characters (2 or 3) as well as a setting where the model determines the number of characters. 
\begin{tcolorbox}[
    colback=gray!10,
    colframe=black,
    width=\columnwidth,
    breakable
]
\ttfamily
You are writing a short, coherent story (<15 sentences) about the entity "\{entity\}". \\
\\
Rules: \\
- The sentence below MUST appear word-for-word, exactly as written in the narrative. \\
- You cannot paraphrase, reword, or change punctuation in that sentence. \\
- Wrap every mention of the entity "{entity}" in <e></e> tags. \\
- The story should naturally include that sentence and expand around it. \\

\noindent\textbf{Character Count Variants:}\\
- \textbf{2 Characters:} The story must have 2 characters \\
- \textbf{3 Characters:} The story must have 3 characters \\
- \textbf{Any Number of Characters:} The number of characters in the story is up to you. \\

MANDATORY SENTENCE (must appear exactly as shown): \\
"\{entity\_sentence\}"
\end{tcolorbox}

\subsection{Narrative Length}
Additionally we experiment with different narrative lengths. The prompt is displayed below. 
\label{app:narrative_length}

\begin{tcolorbox}[
    colback=gray!10,
    colframe=black,
    width=\columnwidth,
    breakable
]
\ttfamily
\textbf{Length Variants:} 5 / 10 / 15 / 20 sentences. \\

You are writing a short, coherent story about the entity "\{entity\}" with the specified length. 

Rules: \\
- The sentence below MUST appear word-for-word, exactly as written in the narrative. \\
- You cannot paraphrase, reword, or change punctuation in that sentence. \\
- Wrap every mention of the entity "\{entity\}" in <e></e> tags. \\
- The story should naturally include that sentence and expand around it. \\
- The number of characters in the story is up to you. \\

MANDATORY SENTENCE (must appear exactly as shown): \\
"\{entity\_sentence\}"
\end{tcolorbox}

\subsection{Final Narrative}
\label{app:final_narrative}
This prompt contains the final instructions used to generate the two different datasets. We name dataset 1 \textbf{Narrative-UFET-Change} and dataset 2 \textbf{Narrative-UFET-Maintain}. 

\begin{tcolorbox}[
    colback=gray!10,
    colframe=black,
    width=\columnwidth,
    breakable
]
\ttfamily
You are writing a short, coherent story that is 10 sentences about the entity "\{entity\}". \\

Rules: \\
- The sentence below MUST appear word-for-word, exactly as written in the narrative. \\
- You cannot paraphrase, reword, or change punctuation in that sentence. \\
- Wrap every mention of the entity "\{entity\}" in <e></e> tags. \\
- The story should naturally include that sentence and expand around it. \\
- The number of characters in the story is up to you. \\
\noindent\textbf{Type Variants:} \\
- \textbf{Change Type:} You must change the type of the entity throughout the story. The number of changes is up to you. \\
- \textbf{Maintain Type:} You must choose one type which represents the entity throughout the story. \\

MANDATORY SENTENCE (must appear exactly as shown): \\
"\{entity\_sentence\}"
\end{tcolorbox}

\section{Evaluation of Narrative Generation Pipeline}

\subsection{Model Selection}\label{app:eval_model_selection}

\begin{figure*}[t]
    \centering
    \includegraphics[width=1\linewidth]{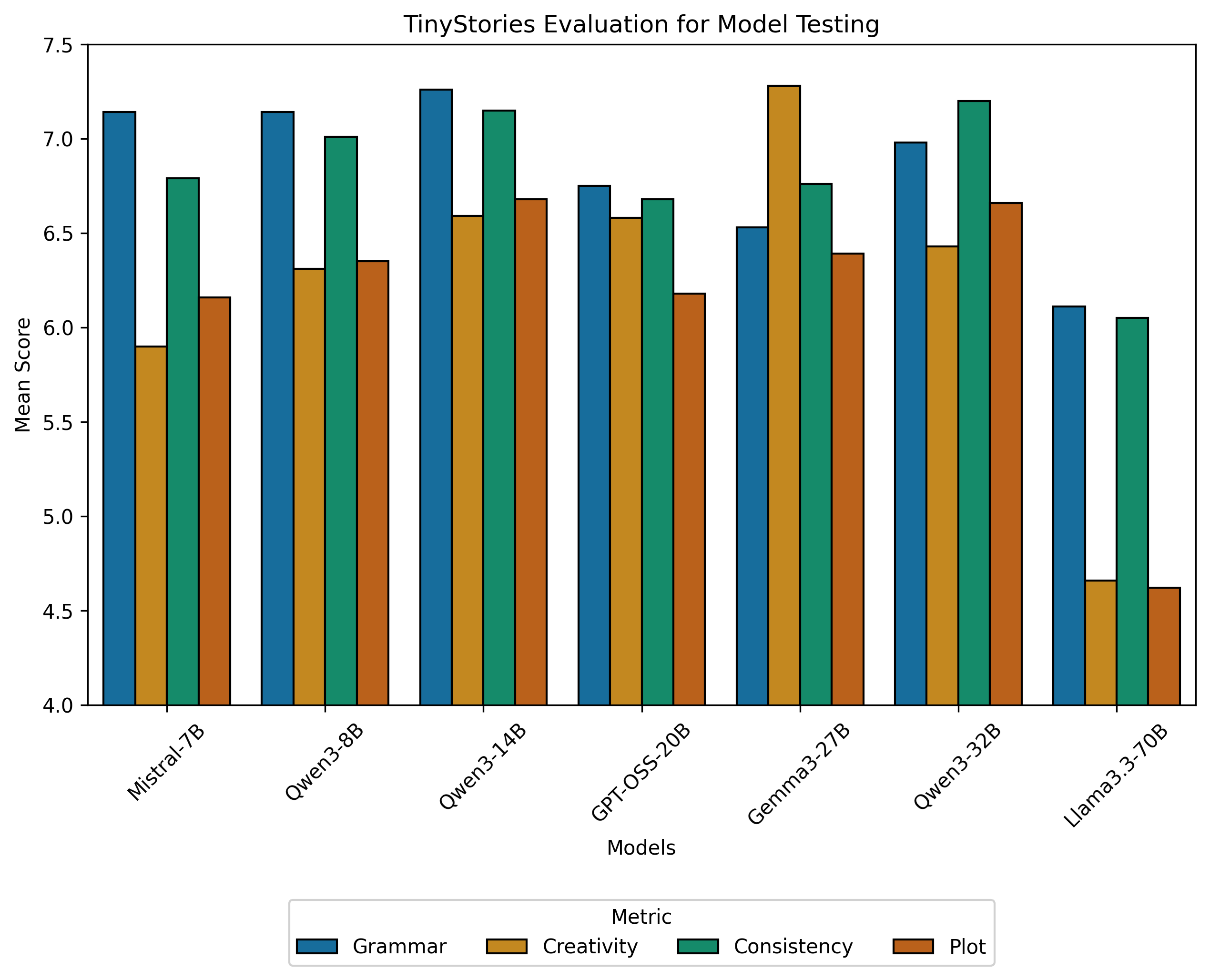}
    \caption{Mean TinyStories narrative quality scores across models for grammar, creativity, consistency, and plot for model testing.}
    \label{fig:tinystories_model_testing}
\end{figure*}

\begin{figure*}[t]
    \centering
    \includegraphics[width=1\linewidth]{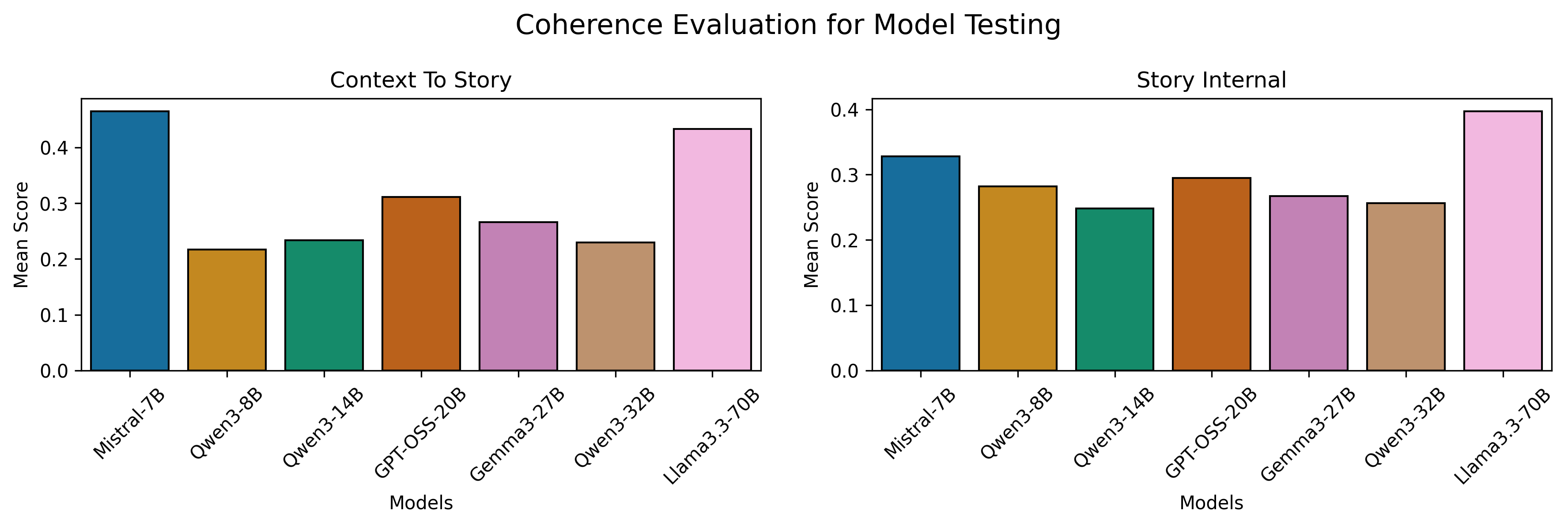}
    \caption{Mean coherence scores across models for model testing.}
    \label{fig:coherence_model_testing}
\end{figure*}

\begin{figure*}[t]
    \centering
    \includegraphics[width=1\linewidth]{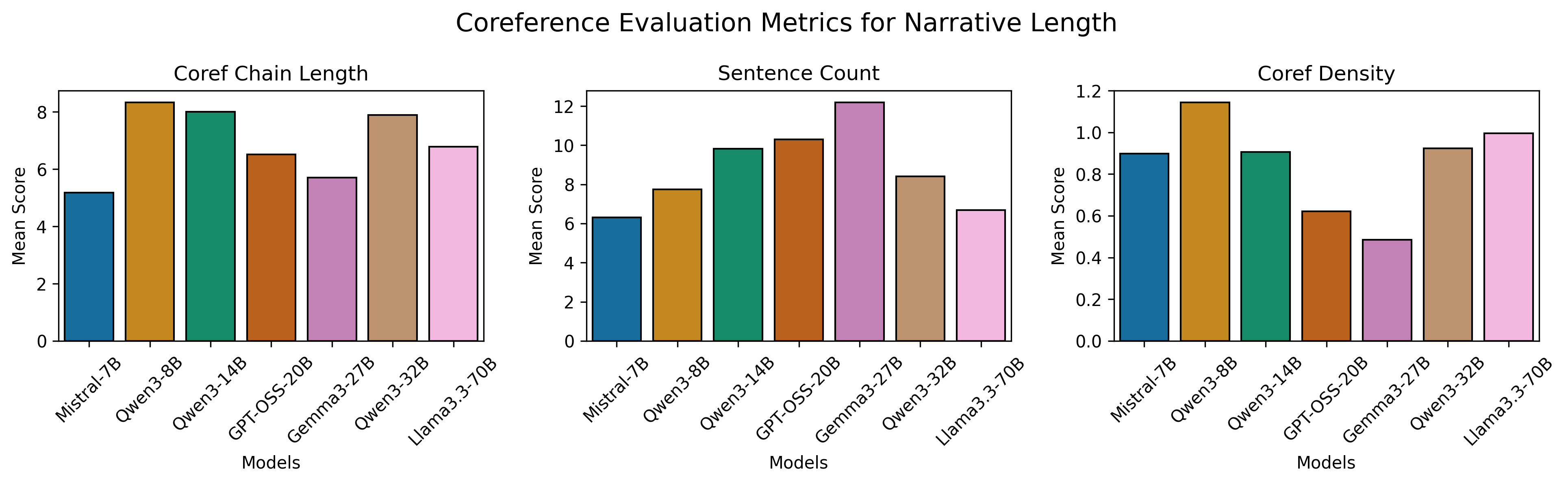}
    \caption{Mean coreference scores across models for model testing.}
    \label{fig:coreference_model_testing}
\end{figure*}

Figures~\ref{fig:tinystories_model_testing}-\ref{fig:coreference_model_testing} present all dimensions of model selection results, including narrative quality, discourse coherence, and coreference chain length.

\subsection{Prompt Design}\label{app:eval_prompt_design}

\subsubsection{Number of Characters}

\begin{figure}[t]
    \centering
    \includegraphics[width=1\columnwidth]{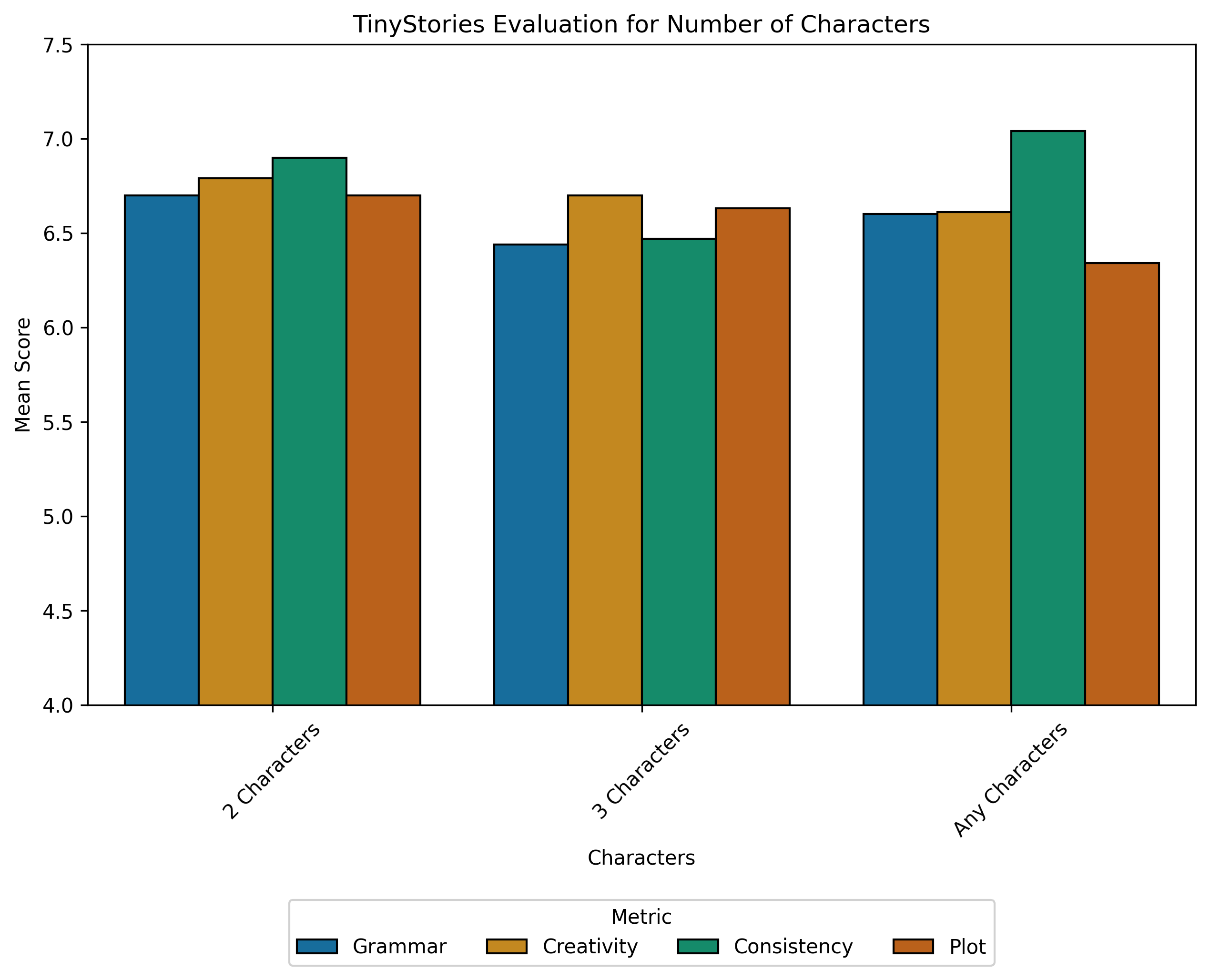}
    \caption{Mean TinyStories narrative quality scores across models for grammar, creativity, consistency, and plot for number of character testing.}
    \label{fig:tinystories_character}
\end{figure}

\begin{figure*}[t]
    \centering
    \includegraphics[width=1\linewidth]{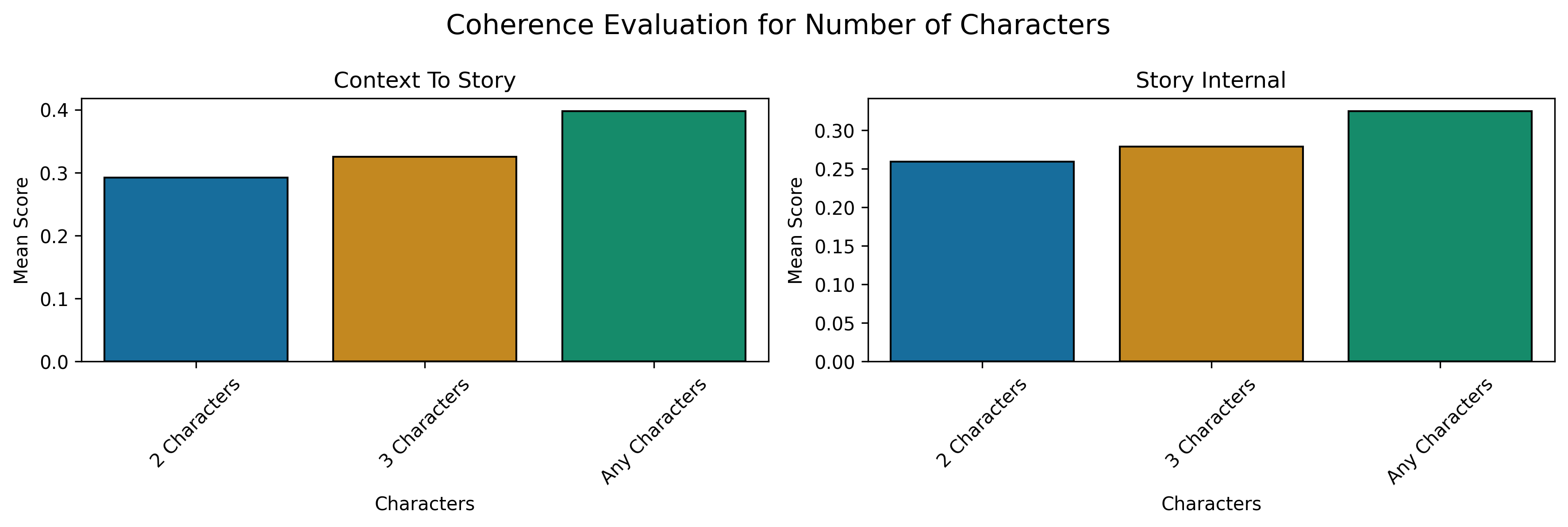}
    \caption{Mean coherence scores for character testing}
    \label{fig:coherence_character}
\end{figure*}

\begin{figure*}[t]
    \centering
    \includegraphics[width=1\linewidth]{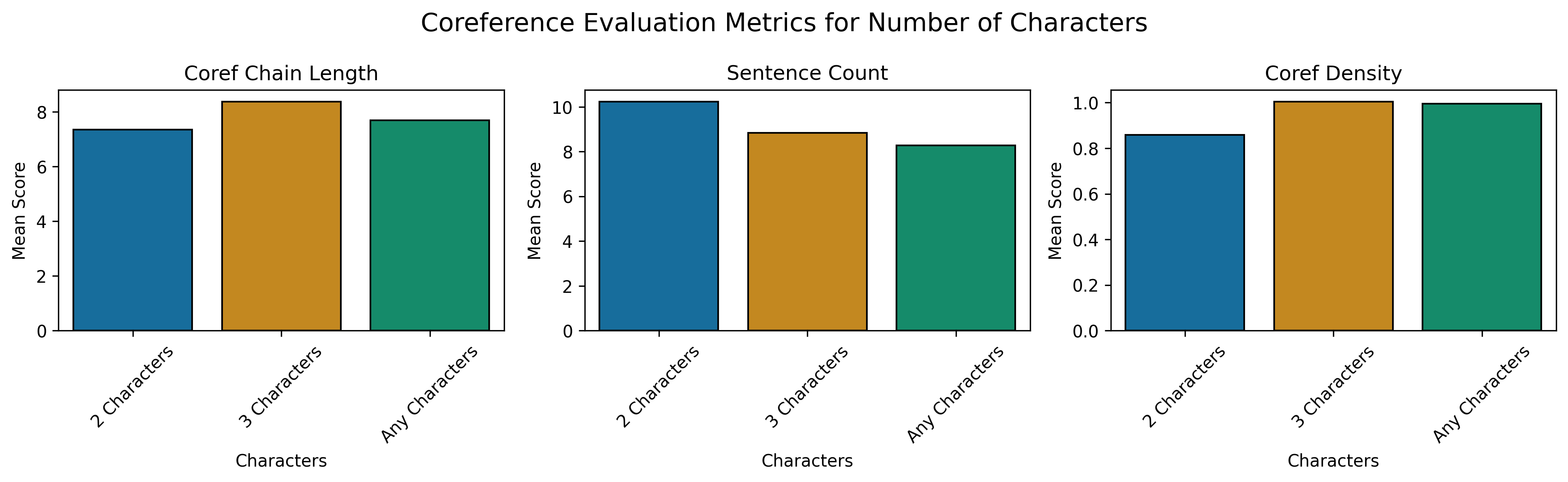}
    \caption{Mean coreference scores for character testing.}
    \label{fig:coreference_character}
\end{figure*}

Figures~\ref{fig:tinystories_character}-\ref{fig:coreference_character} show all evaluation dimensions of the number-of-characters prompt results.


\subsubsection{Narrative Length}

\begin{figure}[t]
    \centering
    \includegraphics[width=1\columnwidth]{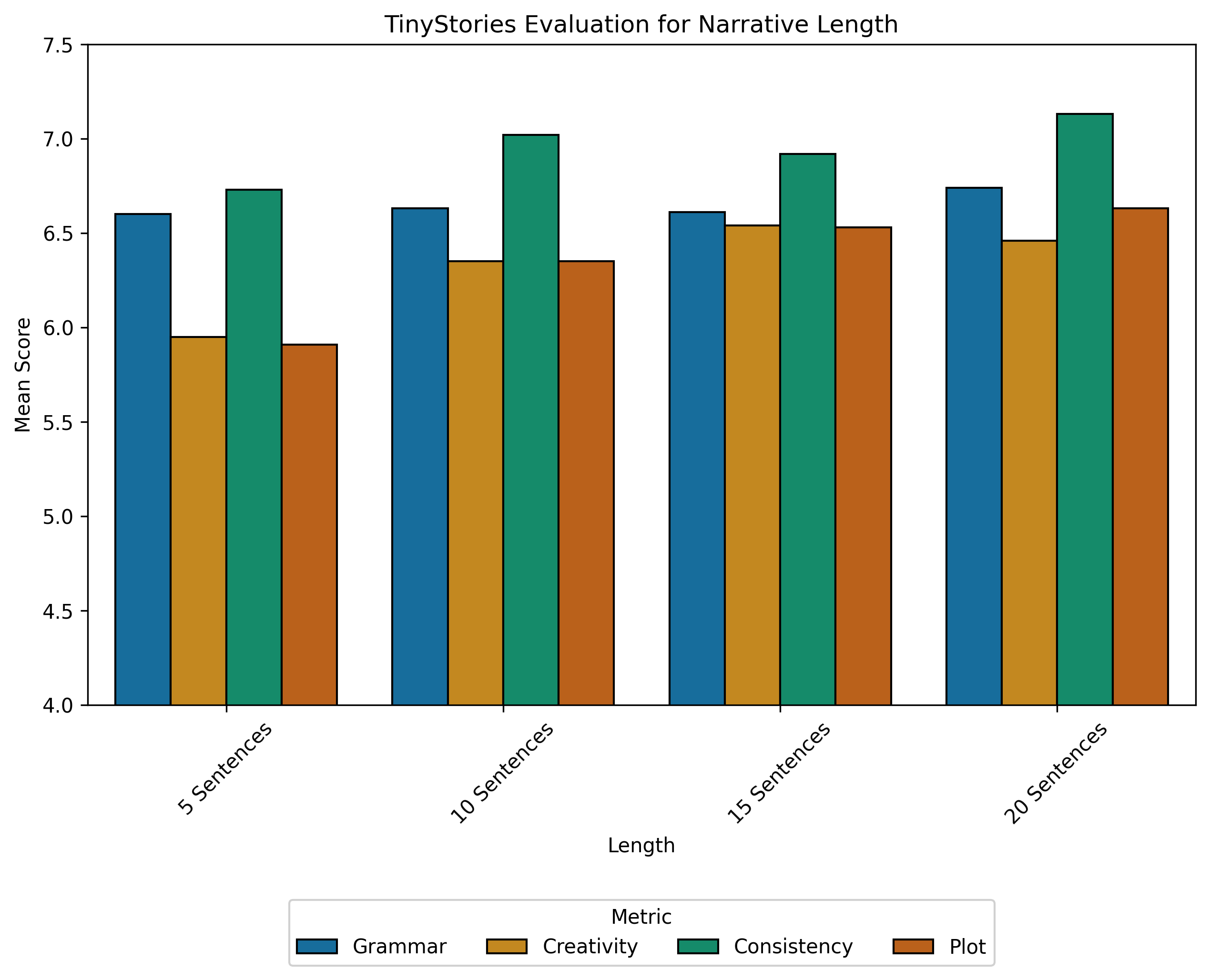}
    \caption{Mean TinyStories narrative quality scores across models for grammar, creativity, consistency, and plot for narrative lengths.}
    \label{fig:tinystories_nar}
\end{figure}

\begin{figure*}[t]
    \centering
    \includegraphics[width=1\linewidth]{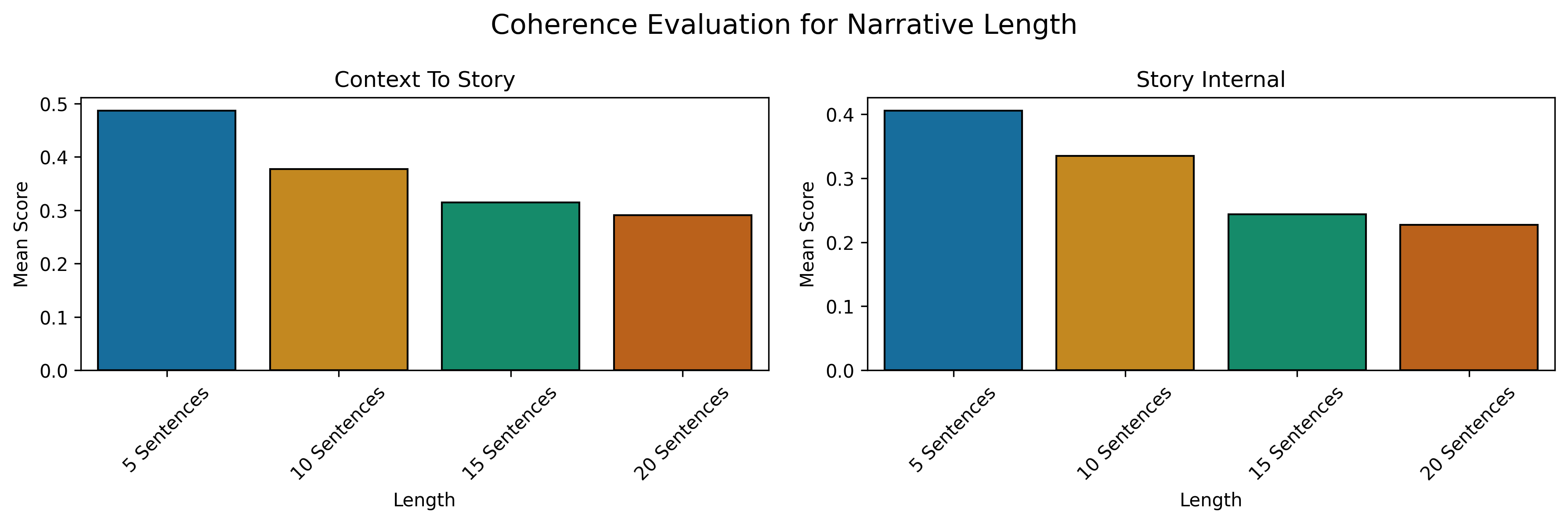}
    \caption{Mean coherence scores for narrative lengths.}
    \label{fig:coherence_nar}
\end{figure*}

\begin{figure}
    \centering
    \includegraphics[width=1\linewidth]{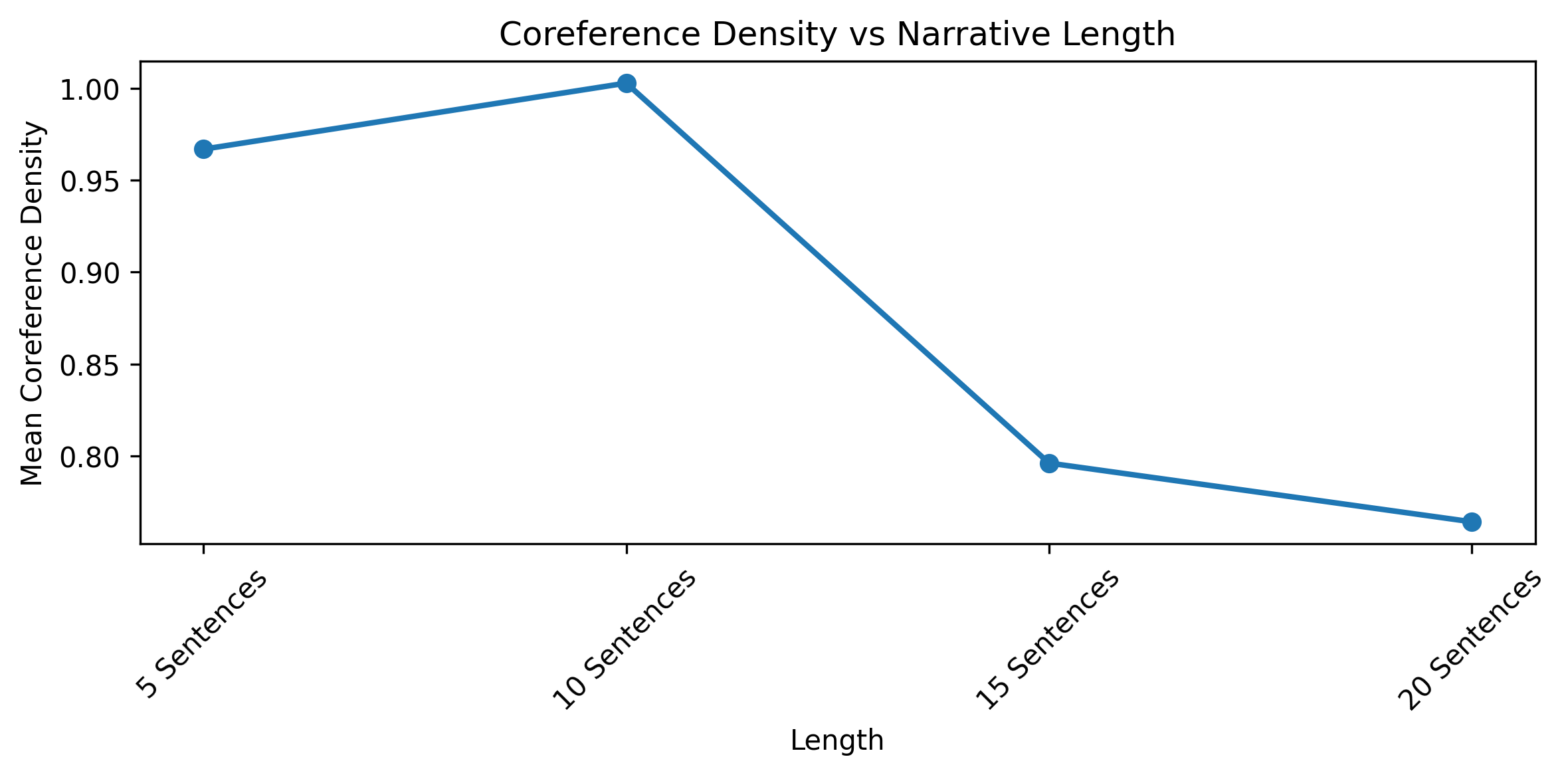}
    \caption{Mean coreference density scores for different narrative lengths.}
    \label{fig:coreference_nar}
\end{figure}

Figures~\ref{fig:tinystories_nar}-\ref{fig:coreference_nar} present all dimensions of the narrative-length results.


\section{Human Evaluation of Narratives}
This section details the human evaluation setup and results for Narrative-UFET.
\label{app:human_eval}

\subsection{Rubric Dimensions and Descriptions}
All rubric dimensions and descriptions are shown in Table~\ref{tab:eval_dimensions}.
\label{app:rubric}

\begin{table}[t]
\centering
\small
\begin{tabularx}{\columnwidth}{lX}
\toprule
\textbf{Dimension} & \textbf{Description} \\
\midrule
Grammar & The narrative is grammatically correct. \\
\midrule
Creativity & The narrative employs creative elements such as genre, narrative perspective, or varied linguistic choices (e.g., metaphors, stylized dialogue). \\
\midrule
Consistency & All mentions of characters, events, and references are internally coherent. \\
\midrule
Plot & The narrative has a clear beginning, middle, and end with logical cause-and-effect; no abrupt transitions, contradictions, or logical flaws. \\
\midrule
Context-to-story & Each sentence in the narrative is semantically related to the original entity-context pair. \\
\midrule
Story-internal & Each sentence is semantically related to its predecessor. \\
\bottomrule
\end{tabularx}
\caption{Dimensions used for human evaluation of narrative quality. Each dimension is rated on a 5-point Likert scale (1 = strongly disagree, 5 = strongly agree).}
\label{tab:eval_dimensions}
\end{table}

\subsection{Annotator Results}
\label{app:annotator_results}

Table~\ref{tab:annotator_results_change} presents the average annotator scores and Gwet (AC2) agreement results for Narrative-UFET-Change and Table~\ref{tab:annotator_results_maintain} presents the same for Narrative-UFET-Maintain. 

\begin{table*}[t]
\centering
\begin{tabularx}{\textwidth}{lXXX}
\hline
 & \textbf{Annotator 1 Average Scores for Narrative-UFET-Change} & \textbf{Annotator 2 Average Scores for Narrative-UFET-Change} & \textbf{Gwet (AC2) Agreement Scores for Narrative-UFET-Change}  \\ \hline
\textbf{Grammar} & 4.51 & 4.44 & 0.8038 \\ 
\textbf{Creativity} & 4.23 & 3.73 & 0.7365 \\ 
\textbf{Consistency} & 4.53 & 4.32 & 0.8855 \\ 
\textbf{Plot} & 3.95 & 4.01 & 0.8601 \\ 
\textbf{Context-to-Story} & 4.53 & 4.44 & 0.8578 \\ 
\textbf{Story-Internal} & 4.72 & 4.65 & 0.8684 \\ \hline
\end{tabularx}
\caption{Annotator \& Gwet (AC2) Results for Narrative-UFET-Change}
\label{tab:annotator_results_change}
\end{table*}

\begin{table*}[t]
\centering
\begin{tabularx}{\textwidth}{lXXX}
\hline
 & \textbf{Annotator 1 Average Scores for Narrative-UFET-Maintain} & \textbf{Annotator 2 Average Scores for Narrative-UFET-Maintain} & \textbf{Gwet (AC2) Agreement Scores for Narrative-UFET-Maintain}  \\ \hline
\textbf{Grammar} & 4.34 & 4.72 & 0.8447 \\ 
\textbf{Creativity} & 4.57 & 3.56 & 0.4489 \\ 
\textbf{Consistency} & 4.42 & 4.42 & 0.7822 \\ 
\textbf{Plot} & 4.43 & 3.99 & 0.7124 \\ 
\textbf{Context-to-Story} & 4.55 & 4.80 & 0.8395 \\ 
\textbf{Story-Internal} & 4.36 & 4.67 & 0.8089 \\ \hline
\end{tabularx}
\caption{Annotator \& Gwet (AC2) Results for Narrative-UFET-Maintain}
\label{tab:annotator_results_maintain}
\end{table*}

\section{Entity-Typing}

This section shows the experimental design and results of 
MLMs and CLMs. 

\subsection{Bin Split Details}
\label{app:bin_split_details}

Following \citet{deshmukh-etal-2025-entities}, we partition the UFET test set into four bins based on entity frequency. Entity frequency scores are calculated for all target UFET entities using the Google Custom Search API where internet search hits are used as a proxy for how often an entity appears in the PLM pretraining data. Entities are then grouped into the four bins based on quartiles. Bin 1 contains the rarest entities and Bin 4 the most frequent. Table~\ref{tab:ufet_bins} shows this bin split. 

To make sure that internet search hits provide a good proxy for estimating realtive entity frequency \citet{deshmukh-etal-2025-entities} validate the proxy by correlating it with real-world datasets known to be used in PLM pretraining, acknowledging that such disclosures are limited to only a few models. Some datasets include BookCorpus \cite{DBLP:journals/corr/ZhuKZSUTF15}, C4 \cite{JMLR:v21:20-074}, and RedPajama \cite{weber2024redpajamaopendatasettraining}. 

Additionally, \citet{deshmukh-etal-2025-entities} recognize that the Internet is constantly evolving and that temporal dynamics could potentially alter the distribution of entities. For this reason, they performed their analysis with API data capped from 2018 to 2024, and found
the results to be consistent over time, with minor changes in correlation coefficients. They also recognize that entity classification can change across large time periods, they compare the correlation between the LM predictions for an entity and the API data capped at 2018 and 2024, they find that results are largely consistent across these two time periods, showing that only 39 entities from the test set (<2\%) changed their bin classifications. For our main results, we rank entities using the
2024 results.

\begin{table}[h]
\centering
\small
\begin{tabular}{p{0.5cm} p{1.2cm} p{3.2cm} p{1.2cm}}
\toprule
\textbf{Bin} & \textbf{\# of Ex.} & \textbf{Representative Entity} & \textbf{Avg. \# Tok.} \\
\midrule
1 & 301 & the Baton Rouge police chief and the serial murder task force & 11.63 \\
\midrule
2 & 301 & Left fielder Carl Crawford & 4.35 \\
\midrule
3 & 300 & The Polish government & 2.67 \\
\midrule
4 & 1095 & the film & 1.18 \\
\bottomrule
\end{tabular}
\caption{Distribution of entities across UFET test bins \cite{deshmukh-etal-2025-entities}}
\label{tab:ufet_bins}
\end{table}

\subsection{MLM Hearst Pattern Details}
\label{app:entity_typing_hearst}

In order to test Narrative-UFET via MLMs, three hearst-like patterns are used, where [MASK] tokens are appended to the entity mention. In this setup, the first entity mentioned in the narrative is replaced by a [MASK], where the following three prompting strategies are used: ‘[MASK] such as {entity},’ ‘{entity} and any other [MASK],’ and ‘{entity} and some other [MASK]’ \cite{deshmukh-etal-2025-entities}. Additionally, we test how many types the model should generate per narrative using the development set, this is represented by the hyperparameter $n$. Values $n=1$ to $n=10$ are tested. The test set is then evaluated using only the hearst-like patterns and hyperparameter $n$.

\subsection{CLM Narrative Prompt for Entity Typing}
\label{app:entity_typing_prompt_nar}
In this experiment, Llama3.3-70B and Qwen3-32B were run in Q4\_K\_M quantized form using Ollama, with the temperature parameter set to 0 for both models \cite{deshmukh-etal-2025-entities}. 15-shot prompting is used with examples taken from the development set. The prompt to generate types auto-regressively is presented below. 
\begin{tcolorbox}[
    colback=gray!10,
    colframe=black,
    width=\columnwidth,
    breakable
]
\ttfamily
\# Entity-Typing Assistant \\
You are a precise entity-typing assistant. \\
\\
Given a narrative in which all entity mentions are wrapped in '<ENT> ... </ENT>' tags, produce only a JSON object whose single key is "predicted\_types". Only predict types for the entity mention wrapped in '<ENT> ... </ENT>' tags. \\
\\
Use the full narrative as context ONLY if it describes that same entity.
Do NOT predict types for the overall narrative topic or other entities. \\

\#\# Guidelines \\
- The value must be a JSON array of strings. \\
- Aim for HIGH RECALL: include every type that is supported by the text, even if it is broad/redundant. \\
- Always include the best broad class when supported
Example: ["person", "organization", "place", "location", "object"]. \\
- Remove duplicates, keep each type concise (ideally a short noun phrase), and use labels consistent to the examples. \\
- If both a broad type and a more specific subtype are valid labels, include BOTH. 
Example: ["college", "community\_college"]. \\
- Do not output any keys other than "predicted\_types". \\
\\
\#\# Input Format \\
- ENTITY\_SENTENCE: The sentence that contains the target entity marked with <ENT> tags (primary evidence).\\
- NARRATIVE: The complete narrative with all target entities clearly mapped in <ENT> tags (secondary evidence).\\
- ENTITY\_MENTION: The target entity mention from the narrative.\\

\#\# Output Format (ONLY JSON)\\
\{\\
  "predicted\_types": ["TypeA", "TypeB", "TypeC", ...]\\
\}
\end{tcolorbox}

\subsection{Entity-Typing Results}
\label{app:entity_typing_results}
\begin{figure}
    \centering
    \includegraphics[width=1\linewidth]{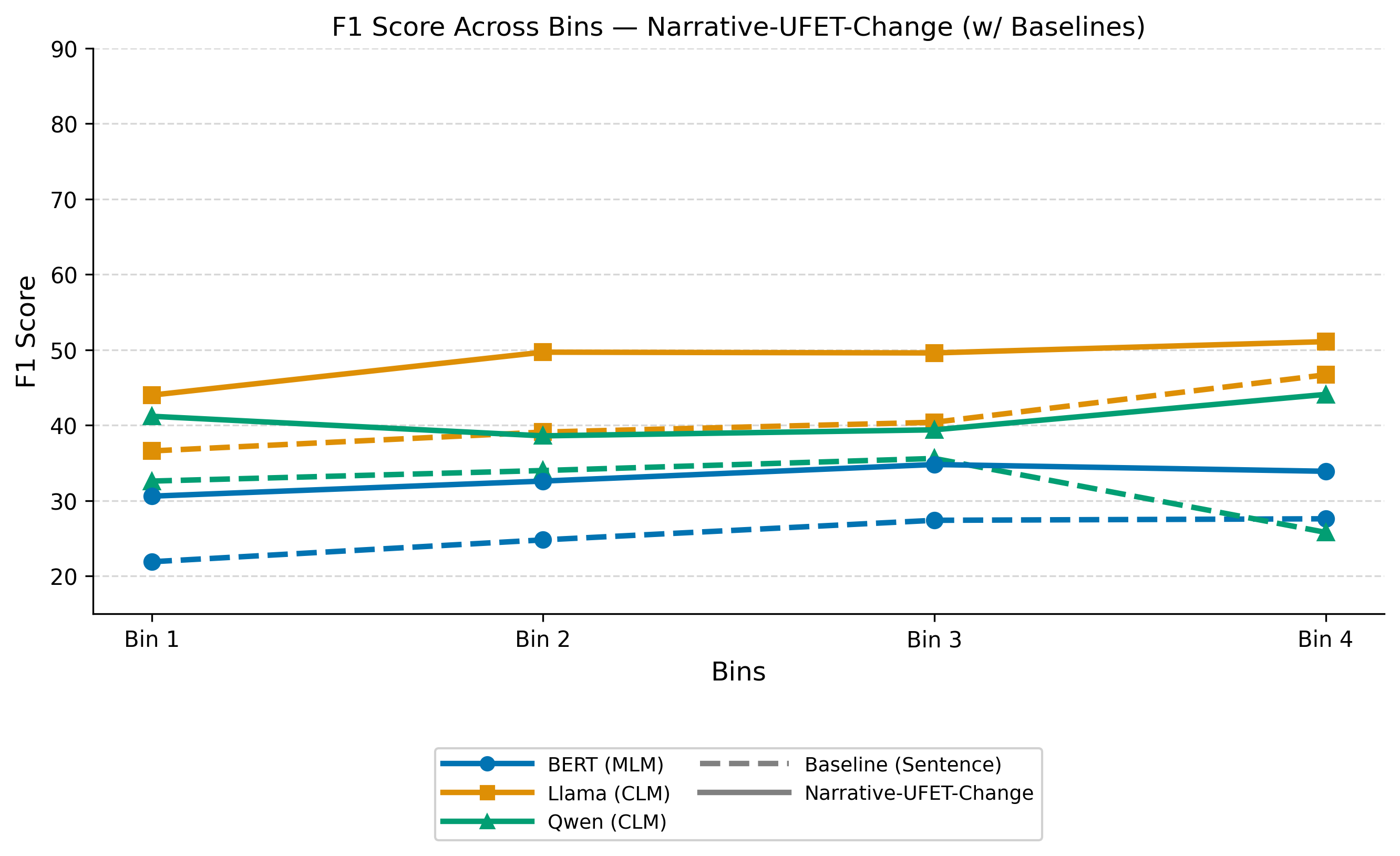}
    \caption{F1 Scores Across Bins. Comparing MLM and CLM performance across Narrative-UFET-Change of Narrative-UFET and the Standard UFET (sentence-level).}
    \label{fig:change-vs-baseline}
\end{figure}

\begin{figure}
    \centering
    \includegraphics[width=1\linewidth]{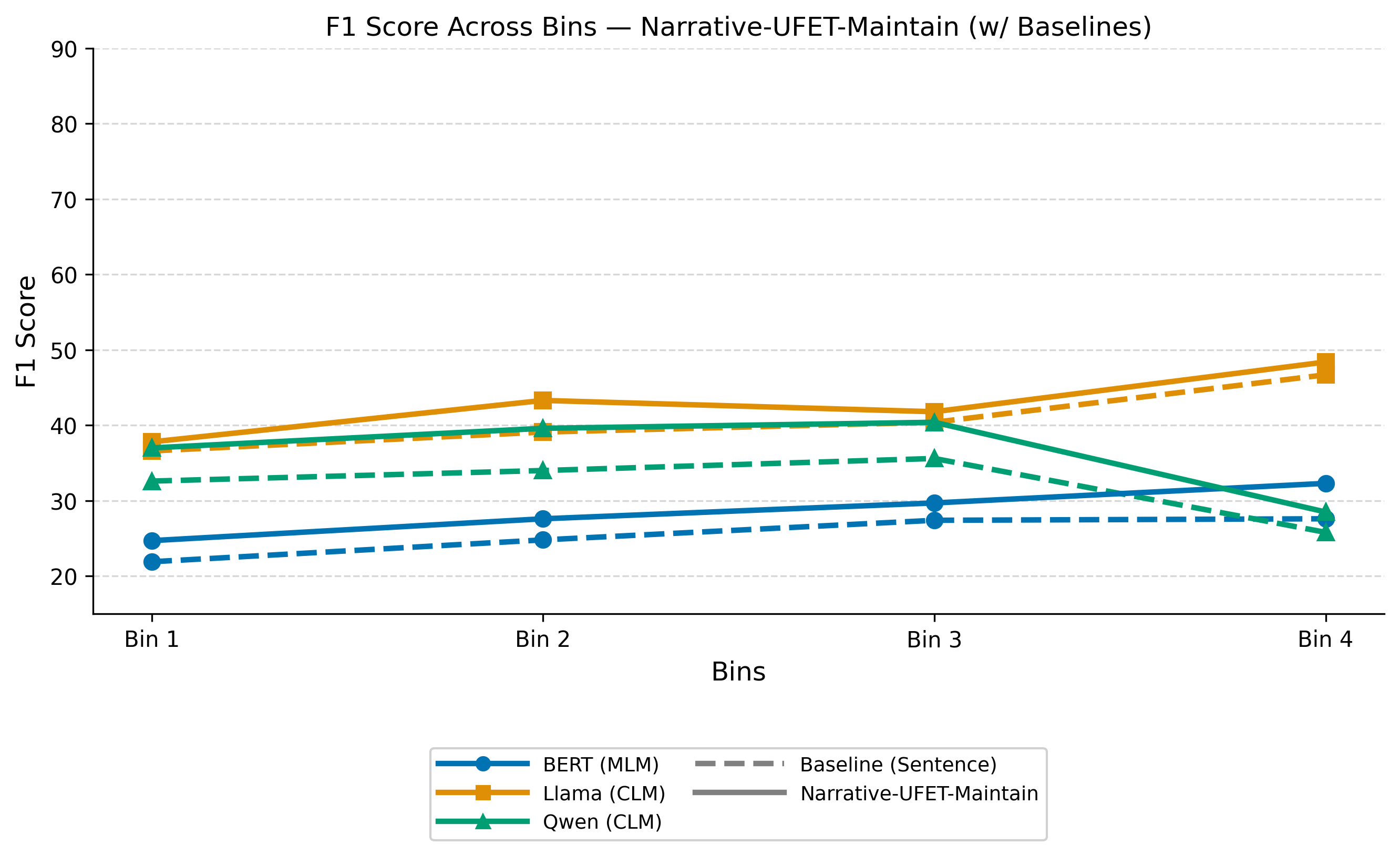}
    \caption{F1 Scores Across Bins. Comparing MLM and CLM performance between Narrative-UFET-Maintain of Narrative-UFET and the Standard UFET (sentence-level).}
    \label{fig:maintain-vs-baseline}
\end{figure}

\begin{figure}
    \centering
    \includegraphics[width=1\linewidth]{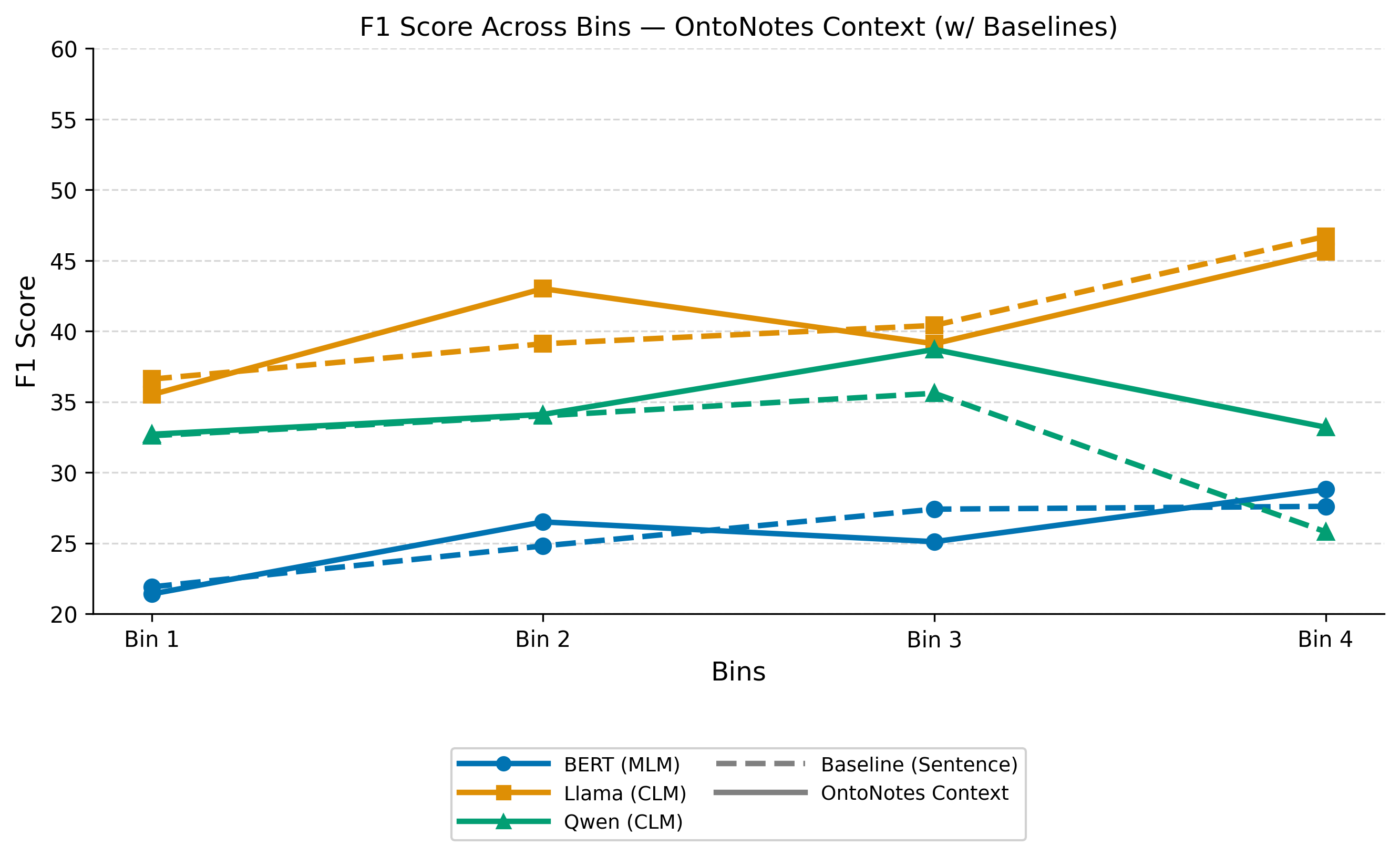}
    \caption{F1 Scores Across Bins. Comparing MLM and CLM performance between original OntoNotes context and the Standard UFET (sentence-level). Scored on the UFET type set.}
    \label{fig:ontonotes-vs-baseline}
\end{figure}

\begin{figure}
    \centering
    \includegraphics[width=1\linewidth]{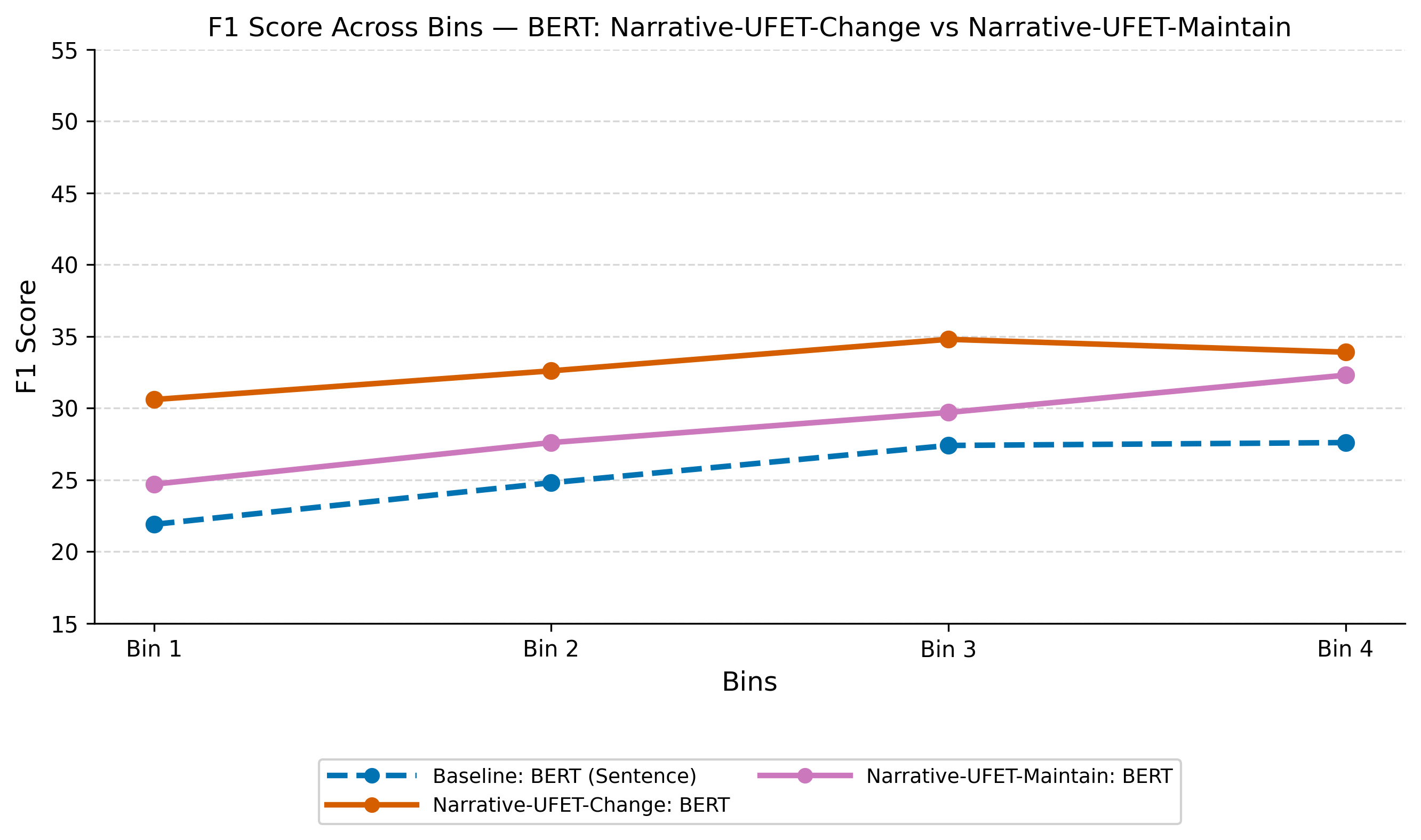}
    \caption{F1 Scores Across Bins. Comparing MLM Bert Model between both datasets of Narrative-UFET and the Standard UFET (sentence-level).}
    \label{fig:bert-vs-datasets}
\end{figure}

\begin{figure}
    \centering
    \includegraphics[width=1\linewidth]{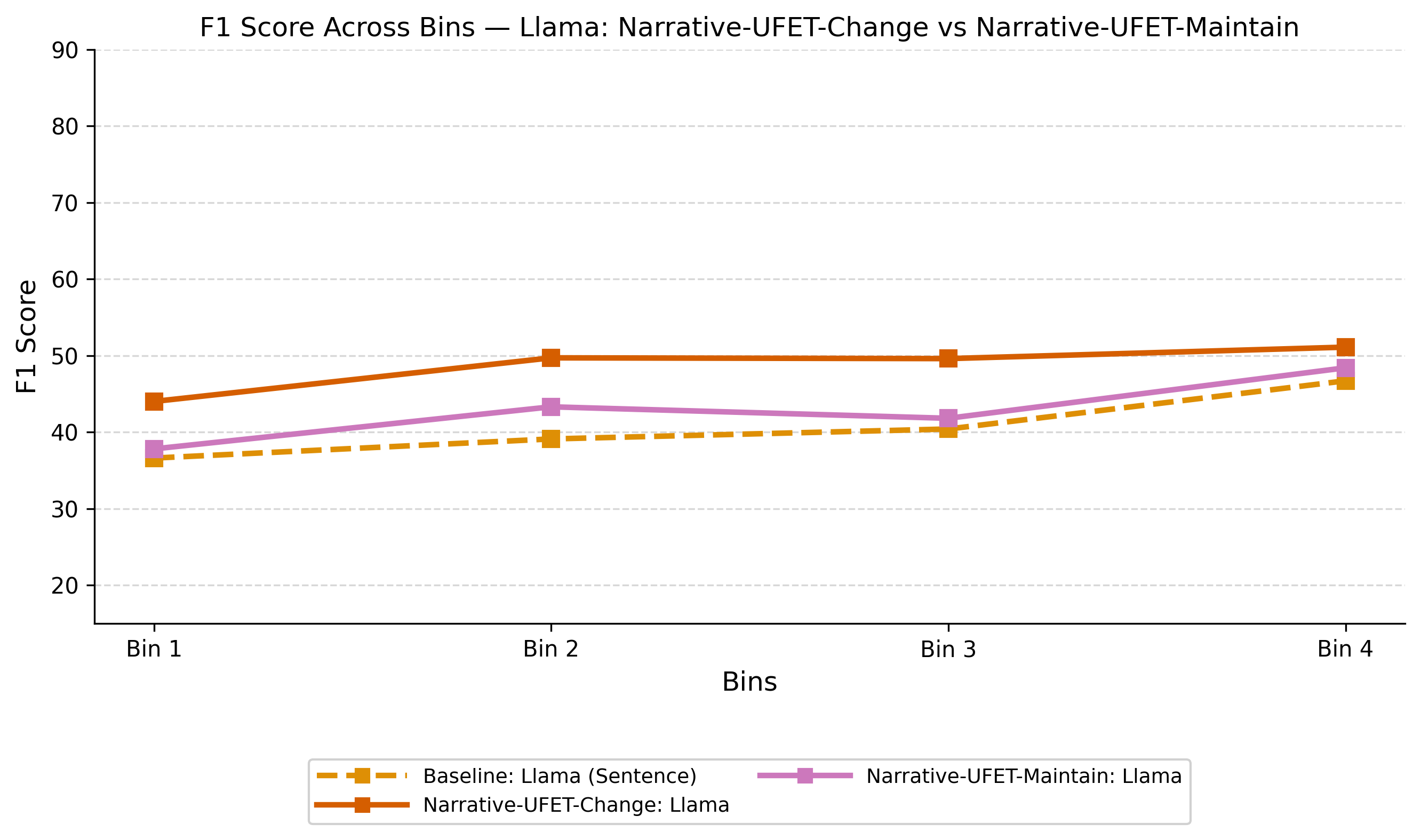}
    \caption{F1 Scores Across Bins. Comparing Llama Model between both datasets of Narrative-UFET and the Standard UFET (sentence-level).}
    \label{fig:llama-vs-datasets}
\end{figure}

\begin{figure}
    \centering
    \includegraphics[width=1\linewidth]{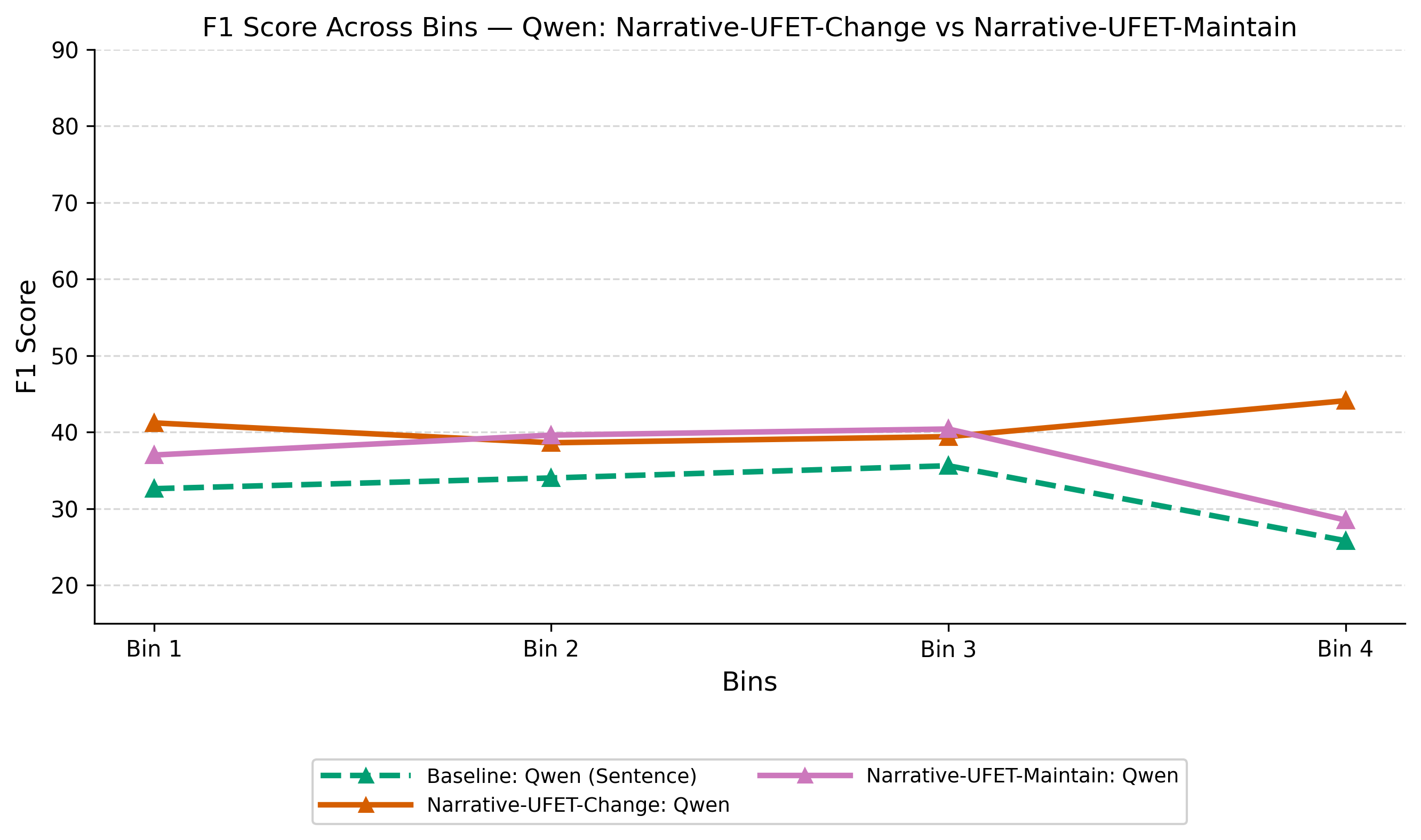}
    \caption{F1 Scores Across Bins. Comparing Qwen Model between both datasets of Narrative-UFET and the Standard UFET (sentence-level).}
    \label{fig:qwen-vs-datasets}
\end{figure}

Tables~\ref{tab:bert-mlm-deshmukh}-\ref{tab:qwen-results-onto} show expanded precision, recall, and F1 scores for the standard and expanded entity-context representation.  Figures~\ref{fig:change-vs-baseline}-\ref{fig:qwen-vs-datasets} show the F1 scores of all models in varying ways. 

\paragraph{Overall Impact of Narrative Context}

Across all three models, providing narrative context yields consistent improvements over Standard-UFET. BERT (Tables~\ref{tab:bert-mlm-deshmukh}--\ref{tab:bert-mlm-maintain}) improves from 26.3 to 33.4 F1 with Narrative-UFET-Change, while Llama (Tables~\ref{tab:llama-results-sent}--\ref{tab:llama-results-maintain}) rises from 42.5 to 49.6, and Qwen (Tables~\ref{tab:qwen-results-sent}--\ref{tab:qwen-results-maintain}) shows the largest absolute gain, jumping from 29.6 to 42.2. The gains are most pronounced at the ultra-fine level, where single-sentence context is least sufficient for this task. BERT's ultra-fine F1 improves from 19.2 to 25.0, and Llama's from 31.6 to 40.2. This confirms that broader narrative context is especially valuable for resolving fine-grained distinctions that a single sentence cannot give. This is a stark contrast from coarse-level performance which was already relatively strong under Standard-UFET and improves more modestly with Narrative-UFET-Maintain.

\paragraph{Narrative-UFET-Change vs Narrative-UFET-Maintain}

Narrative-UFET-Change consistently outperforms Narrative-UFET-Maintain across all models, suggesting that narratives designed for type changes provide stronger typing signals. Bin-level trends show that the hardest examples (Bin 1) benefit substantially from narrative context, for example, Llama's Bin 1 F1 rises from 36.6 to 44.0 under Narrative-UFET-Change (Table~\ref{tab:llama-results-change}), while Bin 4, which is both the easiest and largest subset (1,095 examples), already performs well under Standard-UFET and sees comparatively smaller relative gains. Qwen's Standard-UFET baseline is notably recall-poor (22.8 overall) despite high precision (42.1), a pattern narrative context largely corrects by recovering recall to 34.4 under Narrative-UFET-Change. The one notable exception is Qwen's Bin 4 under Narrative-UFET-Maintain, which drops to 28.5 F1 (Table~\ref{tab:qwen-results-maintain}), below its Standard-UFET Bin 4 score of 25.8 (Table~\ref{tab:qwen-results-sent}),  suggesting that for easier, high-volume examples, poorly matched narrative framing can introduce noise rather than signal.

\paragraph{Validating Narrative-UFET with OntoNotes Context} 

The OntoNotes context gives real-world data of a given UFET sentence. The results of using this context sit consistently between the Standard-UFET baseline and the Narrative-UFET conditions, suggesting that real document context helps but that the structured, entity-focused narratives provide a stronger signal. For BERT (Table~\ref{tab:bert-mlm-onto}), the OntoNotes context yields an overall F1 of 36.1, which exceeds the Standard-UFET baseline of 26.3 but falls short of Narrative-UFET-Change's 33.4. Llama (Table~\ref{tab:llama-results-onto}) follows a similar pattern, reaching 41.7 overall F1 under the additional context compared to 42.5 on Standard-UFET and 49.6 on Narrative-UFET-Change, indicating that raw context alone does not consistently outperform the single-sentence baseline for larger models. Qwen (Table~\ref{tab:qwen-results-onto}) reaches 33.8 overall F1 with the context, above its weak Standard-UFET baseline of 29.6 but well below its 42.2 under Narrative-UFET-Change. Taken together, these results suggest that context quality and framing matter as much as context quantity, but narratives explicitly constructed around the entity mention yield the most reliable gains across models and granularity levels.

\begin{table*}[t]
\centering
\caption{\textbf{BERT MLM (bert-base-uncased) w/ Standard-UFET \cite{deshmukh-etal-2025-entities}}}
\label{tab:bert-mlm-deshmukh}
\begin{tabular*}{\textwidth}{@{\extracolsep{\fill}}lcccccccccccc}
\toprule
& \multicolumn{3}{c}{Overall} & \multicolumn{3}{c}{Coarse} & \multicolumn{3}{c}{Fine} & \multicolumn{3}{c}{Ultra-fine} \\
\cmidrule(lr){2-4} \cmidrule(lr){5-7} \cmidrule(lr){8-10} \cmidrule(lr){11-13}
Subset & P & R & F1 & P & R & F1 & P & R & F1 & P & R & F1 \\
\midrule
Full Test & 23.7 & 29.5 & 26.3 & 64.9 & 43.0 & 51.7 & 34.2 & 40.7 & 37.1 & 16.8 & 22.5 & 19.2 \\
Bin 1     & 19.2 & 25.5 & 21.9 & 59.0 & 41.8 & 48.9 & 22.4 & 34.3 & 27.1 & 14.1 & 19.0 & 16.2 \\
Bin 2     & 22.5 & 27.7 & 24.8 & 57.6 & 34.2 & 43.0 & 31.7 & 43.2 & 36.6 & 17.3 & 23.2 & 19.8 \\
Bin 3     & 24.6 & 31.0 & 27.4 & 60.1 & 41.0 & 48.7 & 34.4 & 42.1 & 37.9 & 18.8 & 25.1 & 21.5 \\
Bin 4     & 25.1 & 30.8 & 27.6 & 69.1 & 45.8 & 55.1 & 38.2 & 41.1 & 39.6 & 16.8 & 22.5 & 19.2 \\
\bottomrule
\end{tabular*}
\end{table*}

\begin{table*}[t]
\centering
\caption{\textbf{BERT MLM (bert-base-uncased) w/ Narrative-UFET-Change}}
\label{tab:bert-mlm-change}
\begin{tabular*}{\textwidth}{@{\extracolsep{\fill}}lcccccccccccc}
\toprule
& \multicolumn{3}{c}{Overall} & \multicolumn{3}{c}{Coarse} & \multicolumn{3}{c}{Fine} & \multicolumn{3}{c}{Ultra-fine} \\
\cmidrule(lr){2-4} \cmidrule(lr){5-7} \cmidrule(lr){8-10} \cmidrule(lr){11-13}
Subset & P & R & F1 & P & R & F1 & P & R & F1 & P & R & F1 \\
\midrule
Full Test & 30.6 & 36.9 & 33.4 & 70.8 & 60.0 & 64.9 & 38.5 & 42.6 & 40.4 & 22.9 & 27.9 & 25.0 \\
Bin 1     & 27.7 & 34.5 & 30.6 & 60.8 & 49.0 & 54.1 & 35.2 & 42.2 & 38.3 & 21.5 & 27.9 & 24.1 \\
Bin 2     & 30.7 & 34.7 & 32.6 & 69.3 & 49.3 & 57.4 & 44.7 & 52.2 & 48.3 & 25.1 & 29.0 & 26.9 \\
Bin 3     & 32.0 & 38.3 & 34.8 & 69.9 & 51.2 & 58.9 & 41.0 & 46.7 & 43.6 & 25.8 & 33.5 & 29.0 \\
Bin 4     & 30.9 & 37.8 & 33.9 & 73.4 & 66.8 & 69.9 & 37.0 & 39.3 & 38.1 & 21.8 & 26.1 & 23.6 \\
\bottomrule
\end{tabular*}
\end{table*}

\begin{table*}[t]
\centering
\caption{\textbf{BERT MLM (bert-base-uncased) w/ Narrative-UFET-Maintain}}
\label{tab:bert-mlm-maintain}
\begin{tabular*}{\textwidth}{@{\extracolsep{\fill}}lcccccccccccc}
\toprule
& \multicolumn{3}{c}{Overall} & \multicolumn{3}{c}{Coarse} & \multicolumn{3}{c}{Fine} & \multicolumn{3}{c}{Ultra-fine} \\
\cmidrule(lr){2-4} \cmidrule(lr){5-7} \cmidrule(lr){8-10} \cmidrule(lr){11-13}
Subset & P & R & F1 & P & R & F1 & P & R & F1 & P & R & F1 \\
\midrule
Full Test & 27.0 & 33.9 & 30.1 & 67.7 & 52.1 & 58.9 & 38.9 & 44.7 & 41.6 & 18.9 & 25.5 & 21.7 \\
Bin 1     & 21.7 & 28.6 & 24.7 & 56.3 & 44.4 & 49.7 & 27.3 & 35.5 & 30.9 & 15.3 & 21.4 & 17.9 \\
Bin 2     & 25.1 & 30.8 & 27.6 & 65.4 & 37.9 & 48.0 & 35.4 & 43.2 & 38.9 & 19.3 & 26.3 & 22.3 \\
Bin 3     & 26.9 & 33.1 & 29.7 & 60.5 & 41.3 & 49.1 & 43.9 & 49.2 & 46.4 & 21.3 & 28.7 & 24.5 \\
Bin 4     & 29.0 & 36.4 & 32.3 & 72.0 & 59.7 & 65.3 & 41.9 & 45.9 & 43.8 & 19.1 & 25.5 & 21.9 \\
\bottomrule
\end{tabular*}
\end{table*}

\begin{table*}[t]
\centering
\caption{\textbf{BERT MLM (bert-base-uncased) w/ OntoNotes Context}}
\label{tab:bert-mlm-onto}
\begin{tabular*}{\textwidth}{@{\extracolsep{\fill}}lcccccccccccc}
\toprule
& \multicolumn{3}{c}{Overall} & \multicolumn{3}{c}{Coarse} & \multicolumn{3}{c}{Fine} & \multicolumn{3}{c}{Ultra-fine} \\
\cmidrule(lr){2-4} \cmidrule(lr){5-7} \cmidrule(lr){8-10} \cmidrule(lr){11-13}
Subset & P & R & F1 & P & R & F1 & P & R & F1 & P & R & F1 \\
\midrule
Full Test & 22.1 & 31.1 & 26.1 & 63.0 & 53.0 & 56.6 & 30.5 & 39.8 & 34.5 & 14.8 & 22.7 & 17.8 \\
Bin 1     & 17.5 & 28.2 & 21.4 & 55.9 & 48.7 & 53.1 & 15.5 & 21.9 & 18.1 & 12.6 & 20.8 & 15.5 \\
Bin 2     & 22.4 & 32.7 & 26.5 & 60.5 & 51.3 & 55.5 & 38.7 & 48.0 & 42.8 & 15.6 & 23.7 & 18.7 \\
Bin 3     & 21.3 & 30.7 & 25.1 & 55.6 & 43.5 & 48.8 & 37.9 & 41.1 & 39.5 & 15.3 & 22.3 & 18.5 \\
Bin 4     & 24.7 & 34.7 & 28.8 & 68.2 & 57.5 & 62.4 & 32.4 & 42.2 & 36.6 & 15.5 & 24.0 & 18.7 \\
\bottomrule
\end{tabular*}
\end{table*}

\begin{table*}[t]
\centering
\caption{\textbf{Llama3.3-70B w/ Standard-UFET}}
\label{tab:llama-results-sent}
\begin{tabular*}{\textwidth}{@{\extracolsep{\fill}}lcccccccccccc}
\toprule
& \multicolumn{3}{c}{Overall} & \multicolumn{3}{c}{Coarse} & \multicolumn{3}{c}{Fine} & \multicolumn{3}{c}{Ultra-fine} \\
\cmidrule(lr){2-4} \cmidrule(lr){5-7} \cmidrule(lr){8-10} \cmidrule(lr){11-13}
Subset & P & R & F1 & P & R & F1 & P & R & F1 & P & R & F1 \\
\midrule
Full Test & 44.6 & 40.6 & 42.5 & 77.9 & 69.8 & 73.6 & 57.9 & 54.7 & 56.2 & 34.7 & 29.0 & 31.6 \\
Bin 1     & 37.8 & 35.4 & 36.6 & 71.2 & 55.9 & 62.6 & 48.3 & 41.7 & 44.8 & 31.2 & 29.0 & 30.1 \\
Bin 2     & 42.6 & 36.1 & 39.1 & 73.5 & 52.7 & 61.4 & 54.1 & 55.4 & 54.7 & 35.6 & 28.6 & 31.7 \\
Bin 3     & 44.8 & 36.8 & 40.4 & 75.3 & 61.5 & 67.7 & 59.4 & 50.2 & 54.4 & 33.9 & 27.1 & 30.1 \\
Bin 4     & 49.4 & 44.2 & 46.7 & 80.5 & 78.9 & 79.7 & 60.5 & 58.5 & 59.5 & 38.5 & 29.6 & 33.5 \\
\bottomrule
\end{tabular*}
\end{table*}

\begin{table*}[t]
\centering
\caption{\textbf{Llama3.3-70B w/ Narrative-UFET-Change}}
\label{tab:llama-results-change}
\begin{tabular*}{\textwidth}{@{\extracolsep{\fill}}lcccccccccccc}
\toprule
& \multicolumn{3}{c}{Overall} & \multicolumn{3}{c}{Coarse} & \multicolumn{3}{c}{Fine} & \multicolumn{3}{c}{Ultra-fine} \\
\cmidrule(lr){2-4} \cmidrule(lr){5-7} \cmidrule(lr){8-10} \cmidrule(lr){11-13}
Subset & P & R & F1 & P & R & F1 & P & R & F1 & P & R & F1 \\
\midrule
Full Test & 45.4 & 55.1 & 49.6 & 77.3 & 85.3 & 81.1 & 59.1 & 64.9 & 61.9 & 37.2 & 43.9 & 40.2 \\
Bin 1     & 39.3 & 50.7 & 44.0 & 73.1 & 70.1 & 71.5 & 50.3 & 57.2 & 53.5 & 32.9 & 44.6 & 37.5 \\
Bin 2     & 45.8 & 54.4 & 49.7 & 80.2 & 75.6 & 77.8 & 59.9 & 70.5 & 64.7 & 38.9 & 46.2 & 42.1 \\
Bin 3     & 45.3 & 55.1 & 49.6 & 72.8 & 76.8 & 74.7 & 62.8 & 67.6 & 65.1 & 38.2 & 45.0 & 41.2 \\
Bin 4     & 46.9 & 56.5 & 51.1 & 78.6 & 92.9 & 85.0 & 60.0 & 64.6 & 62.2 & 37.6 & 42.7 & 39.9 \\
\bottomrule
\end{tabular*}
\end{table*}

\begin{table*}[t]
\centering
\caption{\textbf{Llama3.3-70B w/ Narrative-UFET-Maintain}}
\label{tab:llama-results-maintain}
\begin{tabular*}{\textwidth}{@{\extracolsep{\fill}}lcccccccccccc}
\toprule
& \multicolumn{3}{c}{Overall} & \multicolumn{3}{c}{Coarse} & \multicolumn{3}{c}{Fine} & \multicolumn{3}{c}{Ultra-fine} \\
\cmidrule(lr){2-4} \cmidrule(lr){5-7} \cmidrule(lr){8-10} \cmidrule(lr){11-13}
Subset & P & R & F1 & P & R & F1 & P & R & F1 & P & R & F1 \\
\midrule
Full Test & 45.3 & 44.8 & 45.1 & 74.5 & 78.2 & 76.3 & 56.7 & 55.9 & 56.3 & 36.1 & 32.1 & 34.0 \\
Bin 1     & 36.3 & 39.3 & 37.8 & 67.0 & 63.2 & 65.1 & 49.3 & 42.4 & 45.6 & 29.1 & 30.8 & 29.9 \\
Bin 2     & 43.3 & 43.3 & 43.3 & 72.7 & 65.8 & 69.1 & 53.2 & 59.7 & 56.3 & 35.4 & 33.7 & 34.6 \\
Bin 3     & 42.9 & 40.7 & 41.8 & 71.7 & 70.6 & 71.1 & 58.0 & 53.9 & 55.9 & 33.7 & 30.0 & 31.7 \\
Bin 4     & 49.0 & 47.8 & 48.4 & 77.0 & 86.1 & 81.3 & 58.6 & 58.5 & 58.5 & 38.9 & 32.6 & 35.5 \\
\bottomrule
\end{tabular*}
\end{table*}

\begin{table*}[t]
\centering
\caption{\textbf{Llama3.3-70B w/ OntoNotes Context}}
\label{tab:llama-results-onto}
\begin{tabular*}{\textwidth}{@{\extracolsep{\fill}}lcccccccccccc}
\toprule
& \multicolumn{3}{c}{Overall} & \multicolumn{3}{c}{Coarse} & \multicolumn{3}{c}{Fine} & \multicolumn{3}{c}{Ultra-fine} \\
\cmidrule(lr){2-4} \cmidrule(lr){5-7} \cmidrule(lr){8-10} \cmidrule(lr){11-13}
Subset & P & R & F1 & P & R & F1 & P & R & F1 & P & R & F1 \\
\midrule
Full Test & 37.7 & 47.2 & 41.7 & 71.3 & 75.3 & 73.2 & 52.0 & 56.0 & 53.9 & 28.6 & 34.7 & 31.2 \\
Bin 1     & 30.2 & 44.5 & 35.5 & 68.3 & 66.2 & 67.3 & 41.2 & 43.3 & 42.3 & 24.8 & 34.5 & 28.4 \\
Bin 2     & 39.0 & 48.4 & 43.0 & 77.6 & 86.8 & 81.9 & 71.5 & 75.9 & 73.6 & 26.8 & 34.6 & 30.0 \\
Bin 3     & 35.8 & 43.4 & 39.1 & 60.2 & 56.9 & 58.5 & 45.7 & 55.5 & 50.0 & 30.2 & 32.6 & 31.3 \\
Bin 4     & 42.0 & 50.1 & 45.6 & 74.8 & 83.8 & 79.0 & 50.5 & 51.8 & 51.2 & 30.7 & 35.8 & 33.0 \\
\bottomrule
\end{tabular*}
\end{table*}

\begin{table*}[t]
\centering
\caption{\textbf{Qwen3-32B w/ Standard-UFET}}
\label{tab:qwen-results-sent}
\begin{tabular*}{\textwidth}{@{\extracolsep{\fill}}lcccccccccccc}
\toprule
& \multicolumn{3}{c}{Overall} & \multicolumn{3}{c}{Coarse} & \multicolumn{3}{c}{Fine} & \multicolumn{3}{c}{Ultra-fine} \\
\cmidrule(lr){2-4} \cmidrule(lr){5-7} \cmidrule(lr){8-10} \cmidrule(lr){11-13}
Subset & P & R & F1 & P & R & F1 & P & R & F1 & P & R & F1 \\
\midrule
Full Test & 42.1 & 22.8 & 29.6 & 63.7 & 44.2 & 52.1 & 43.0 & 30.1 & 35.4 & 31.3 & 14.4 & 19.7 \\
Bin 1     & 47.1 & 24.9 & 32.6 & 74.4 & 41.8 & 53.5 & 50.0 & 32.5 & 39.4 & 36.1 & 19.1 & 25.0 \\
Bin 2     & 52.8 & 25.1 & 34.0 & 77.7 & 40.0 & 52.8 & 64.1 & 39.4 & 48.8 & 43.6 & 17.8 & 25.3 \\
Bin 3     & 54.2 & 26.5 & 35.6 & 73.0 & 43.0 & 54.1 & 69.1 & 40.7 & 51.3 & 45.6 & 19.1 & 26.9 \\
Bin 4     & 34.5 & 20.6 & 25.8 & 57.8 & 45.9 & 51.2 & 32.0 & 24.6 & 27.8 & 22.5 & 10.9 & 14.7 \\
\bottomrule
\end{tabular*}
\end{table*}

\begin{table*}[t]
\centering
\caption{\textbf{Qwen3-32B w/ Narrative-UFET-Change}}
\label{tab:qwen-results-change}
\begin{tabular*}{\textwidth}{@{\extracolsep{\fill}}lcccccccccccc}
\toprule
& \multicolumn{3}{c}{Overall} & \multicolumn{3}{c}{Coarse} & \multicolumn{3}{c}{Fine} & \multicolumn{3}{c}{Ultra-fine} \\
\cmidrule(lr){2-4} \cmidrule(lr){5-7} \cmidrule(lr){8-10} \cmidrule(lr){11-13}
Subset & P & R & F1 & P & R & F1 & P & R & F1 & P & R & F1 \\
\midrule
Full Test & 55.4 & 34.4 & 42.2 & 81.3 & 59.1 & 68.3 & 65.0 & 45.3 & 53.2 & 44.0 & 25.1 & 31.5 \\
Bin 1     & 49.1 & 35.8 & 41.2 & 74.7 & 50.8 & 60.3 & 57.0 & 41.5 & 47.9 & 42.2 & 30.0 & 34.9 \\
Bin 2     & 51.5 & 31.3 & 38.6 & 77.6 & 50.1 & 60.6 & 57.0 & 38.6 & 45.8 & 43.7 & 24.8 & 31.2 \\
Bin 3     & 50.4 & 32.7 & 39.4 & 75.1 & 54.0 & 62.7 & 61.8 & 40.4 & 48.6 & 39.2 & 24.1 & 29.5 \\
Bin 4     & 59.9 & 35.4 & 44.1 & 84.9 & 64.1 & 73.0 & 69.3 & 49.0 & 57.3 & 46.1 & 24.1 & 31.1 \\
\bottomrule
\end{tabular*}
\end{table*}

\begin{table*}[t]
\centering
\caption{\textbf{Qwen3-32B w/ Narrative-UFET-Maintain}}
\label{tab:qwen-results-maintain}
\begin{tabular*}{\textwidth}{@{\extracolsep{\fill}}lcccccccccccc}
\toprule
& \multicolumn{3}{c}{Overall} & \multicolumn{3}{c}{Coarse} & \multicolumn{3}{c}{Fine} & \multicolumn{3}{c}{Ultra-fine} \\
\cmidrule(lr){2-4} \cmidrule(lr){5-7} \cmidrule(lr){8-10} \cmidrule(lr){11-13}
Subset & P & R & F1 & P & R & F1 & P & R & F1 & P & R & F1 \\
\midrule
Full Test & 42.7 & 27.2 & 33.2 & 66.5 & 51.8 & 58.2 & 45.5 & 32.9 & 38.2 & 33.1 & 18.3 & 23.6 \\
Bin 1     & 47.6 & 30.2 & 37.0 & 72.4 & 51.7 & 60.3 & 58.6 & 33.1 & 42.3 & 38.2 & 23.5 & 29.1 \\
Bin 2     & 53.4 & 31.4 & 39.6 & 77.4 & 50.9 & 61.4 & 64.0 & 42.4 & 51.0 & 47.1 & 24.7 & 32.4 \\
Bin 3     & 53.3 & 32.5 & 40.4 & 78.1 & 59.0 & 67.2 & 66.4 & 43.8 & 52.8 & 40.7 & 22.4 & 28.9 \\
Bin 4     & 35.5 & 23.8 & 28.5 & 60.8 & 50.4 & 55.1 & 35.0 & 27.6 & 30.9 & 25.5 & 13.9 & 18.0 \\
\bottomrule
\end{tabular*}
\end{table*}

\begin{table*}[t]
\centering
\caption{\textbf{Qwen3-32B w/ OntoNotes Context}}
\label{tab:qwen-results-onto}
\begin{tabular*}{\textwidth}{@{\extracolsep{\fill}}lcccccccccccc}
\toprule
& \multicolumn{3}{c}{Overall} & \multicolumn{3}{c}{Coarse} & \multicolumn{3}{c}{Fine} & \multicolumn{3}{c}{Ultra-fine} \\
\cmidrule(lr){2-4} \cmidrule(lr){5-7} \cmidrule(lr){8-10} \cmidrule(lr){11-13}
Subset & P & R & F1 & P & R & F1 & P & R & F1 & P & R & F1 \\
\midrule
Full Test & 42.4 & 28.3 & 33.8 & 69.3 & 50.6 & 58.3 & 57.5 & 43.8 & 49.6 & 29.6 & 17.3 & 21.4 \\
Bin 1     & 34.5 & 31.1 & 32.7 & 71.1 & 48.8 & 57.4 & 44.4 & 39.2 & 41.6 & 25.5 & 20.7 & 22.7 \\
Bin 2     & 44.7 & 28.4 & 34.1 & 76.7 & 60.0 & 67.1 & 62.8 & 50.9 & 56.1 & 36.0 & 20.0 & 24.5 \\
Bin 3     & 45.8 & 23.9 & 38.7 & 65.6 & 49.0 & 55.9 & 55.0 & 46.4 & 50.2 & 34.7 & 23.3 & 27.3 \\
Bin 4     & 44.9 & 26.3 & 33.2 & 67.8 & 50.2 & 57.7 & 65.2 & 42.6 & 51.5 & 29.5 & 13.4 & 18.5 \\
\bottomrule
\end{tabular*}
\end{table*}

\end{document}

%% file: latex/1_intro.tex
\section{Introduction}


Ultra-fine entity typing (UFET) is the task of assigning highly specific types to entity mentions based on the context in which they appear~\cite{choi-etal-2018-ultra}. Unlike coarse-grained typing, which selects from a small set of broad categories such as \textit{person} or \textit{organization}, UFET aims to capture contextually relevant distinctions: in the sentence \textit{``In his next note on Baidu, he wrote that the company is trading above its fair value,''}, the pronoun ``he'' could be typed coarsely as a \textit{person}, but the surrounding context supports more specific types, such as \textit{writer}, \textit{editor}, or \textit{analyst}. Identifying these fine-grained types is valuable for a wide range of downstream tasks, including coreference resolution~\cite{onoe-durrett-2020-interpretable}, entity linking~\cite{ling-etal-2015-design}, relation extraction \cite{koch-etal-2014-type}, knowledge graph completion \cite{li-etal-2024-integration} and multi-modal entity recognition and grounding \cite{wang-etal-2024-granular, li-etal-2024-llms}.

A central challenge for UFET is the long tail of the entity distribution. Pretrained language models (PLMs), which most current approaches build on~\cite{dai-etal-2021-ultra,li-etal-2023-ultra,deshmukh-etal-2025-entities}, perform well on entities that appear frequently in their pretraining corpora but degrade sharply on rare ones~\cite{deshmukh-etal-2025-entities}. This is precisely the setting where fine-grained disambiguation is most needed. We hypothesize that part of this gap stems from a reliance on sentence-level context. In realistic settings such as articles or books, the evidence needed to disambiguate fine-grained types is rarely contained in a single sentence. Instead, it must be pieced together from cues distributed across the surrounding narrative. Testing this hypothesis has been difficult, however, because all existing annotated resources for fine and ultra-fine entity typing are sentence-level~\cite{Riedel2010ModelingRA,Gillick2014ContextDependentFE,choi-etal-2018-ultra,Ling_Weld_2021,ding-etal-2021-nerd}.

To address this gap we construct Narrative-UFET, an extended version of the UFET dataset~\cite{choi-etal-2018-ultra} in which each entity-sentence pair is paired with a short, automatically generated narrative built around the target entity. We choose to synthesize narratives rather than retrieve them from real corpora because synthesis lets us control specific properties of the discourse, holding everything else fixed. This control is the core methodological move of the paper, as it lets us isolate the effect of a single discourse property on typing performance. As a case study, we construct two variants of Narrative-UFET that differ in one such property: whether the entity's type is held constant across the narrative (\textit{Maintain}) or shifts across it (\textit{Change}). We use this contrast to test whether type variation across discourse is a useful signal for ultra-fine typing.

We validate the quality of the generated narratives through automated metrics and human evaluation. Evaluating both masked and causal PLMs on the UFET task, we find that narrative context yields consistent improvements on long-tail types over sentence-level baselines, with the \textit{Change} variant providing the stronger signal. A comparison against naturally occurring contexts shows that our synthetic narratives yield stronger gains than real text alone, indicating that controlled discourse construction can surface signal that real text leaves implicit. At the same time, substantial room for improvement remains, suggesting that progress on long-tail entity typing will require advances in both discourse-aware modeling and narrative construction beyond the single dimension studied here.

Our contributions are: (i) \textbf{Narrative-UFET}, a controlled narrative-level extension of UFET with two variants (\textit{Maintain} and \textit{Change}); (ii) evidence that narrative context improves typing on long-tail entities, with \textit{Change} providing the strongest signal; and (iii) a comparison against naturally occurring contexts showing that synthetic narratives surface signal that real text leaves implicit.

%% file: latex/2_related.tex
\section{Related Work}


\textbf{Narrative Generation and Evaluation.}
With the rise of powerful generative models, large language models (LLMs) are increasingly used to generate high-quality short and long narratives that exhibit more nuanced plot development and stylistic variation \cite{goldfarb-tarrant-etal-2020-content, yang-etal-2022-re3, harel-canada-etal-2024-measuring}. Additionally, recent work has explored the use of LLMs in assessing the quality of narratives. This is a complex task as quality is multi-faceted and at times subjective. Prior work has evaluated grammar as language fluency \cite{naismith-etal-2023-automated}, creativity and originality as indicators of narrative novelty, and consistency in plot and character progression to assess story coherence over time \cite{chhun-etal-2022-human, tian-etal-2024-large-language}. Recent research also emphasizes the importance of prompt engineering in guiding LLMs to produce more coherent and contextually appropriate narratives \cite{TANG2024e34262}. Although prior work has advanced narrative generation and evaluation, it remains unclear how well models can expand an entity-context pair into a complete narrative that is coherent and remains reliable when evaluated on the entity-typing task. To construct Narrative-UFET we conduct model and prompt testing to generate narratives. We then verify the validity of Narrative-UFET through automated and human evaluation. 

\textbf{UFET and Pre-trained Language Models.} 
UFET was introduced by \citet{choi-etal-2018-ultra}, who proposed the task of predicting free-form type labels for a target entity mention given the sentence in which it appears. Subsequent work leveraged PLMs to improve entity typing performance. Many of these approaches frame UFET as a masked language modeling problem. \cite{dai-etal-2021-ultra} use Hearst patterns with [MASK] tokens and BERT to predict entity types. Similarly, \citet{ijcai2022p599} generate ultra-fine type predictions by appending an entity mention and a [MASK] token to the input sentence. \citet{deshmukh-etal-2025-entities} expand on this work by exploring how such approaches perform on infrequent or rare entities, showing that PLMs struggle with long-tail entity typing unless additional knowledge about rare entities is incorporated, and even then the explored knowledge injection strategies proved insufficient. Motivated by these findings, we use Narrative-UFET to investigate whether richer discourse context can help PLMs better predict long-tail types, evaluating both masked and causal PLMs.


%% file: latex/3_dataset.tex
\section{Narrative-UFET}
\label{sec:3}

In this section, we describe the construction and evaluation of Narrative-UFET. We build on the crowd-annotated portion of the UFET dataset~\cite{choi-etal-2018-ultra}, which contains 5,994 entity mentions paired with their surrounding sentence context and a set of human-annotated ultra-fine types (hereafter, gold types), evenly divided into training, development, and test splits. 

\subsection{Narrative Generation Pipeline}~\label{sec:gen_pipeline}
For each entity-sentence pair in the UFET dataset, we generated a short, self-contained narrative that embeds the original sentence verbatim while building a coherent narrative chain around the target entity. An example narrative is shown in Appendix~\ref{app:narrative_example}. The generation pipeline involved three stages: model selection, prompt design, and the generation of final datasets.

\textbf{Model Selection.} We generated narratives for the first 100 instances from the development set by instructing seven different models\footnote{GPT-OSS-20B, Llama3.3-70B, Gemma3-27B, Qwen3-8B, Qwen3-14B, Qwen3-32B, and Mistral-7B} to produce short, coherent stories around each target entity, requiring that the original UFET sentence appear verbatim (prompt and example narrative in Appendices~\ref{app:model_testing_prompt} and \ref{app:narrative_example}). 

We test quality across three dimensions: (1) \textit{Narrative Quality}, using the TinyStories framework \cite{eldan2023tinystoriessmalllanguagemodels}, which scores grammar, creativity, consistency, and plot on a 1–10 scale. Prompt and other details are shown in Appendix~\ref{app:tiny_stories_eval_framework}; (2) \textit{Discourse Coherence}, measured as both context-to-story alignment (how semantically related each sentence in the narrative is to the original sentence
) and story-internal coherence (how related each sentence is to its prior sentence). Appendix~\ref{app:discourse_coherence} shows implementation details; and (3) \textit{Coreference Density}, which is split into two categories, coreference chain length which is computed as the total number of target entity mentions across the narrative, where longer chains indicate richer entity-centered context (as there is now more information available for that entity), and coreference density which uses the narrative sentence count to normalize the coreference chain length. This is a helpful metric for when model generations do not have an identical number of sentences. All metrics are given equal importance. 

Additionally, a qualitative analysis is done to see if human judgment matches the scores. This analysis is conducted by a single human annotator without specific guidelines, relying on personal judgment to evaluate the generated narratives.

We found that models were inconsistent across evaluation dimensions. The Gemma3-27B model only performed well with \textit{Narrative Quality}, while the Mistral-7B model only performed well in \textit{Discourse Coherence}. 
Qwen3 family models were consistently strong in all three dimensions. The qualitative review revealed that Mistral-7B, Llama3.3-70B, and Qwen3-8B frequently did not include the original sentence verbatim, while GPT-OSS-20B and Gemma3-27B produced repetitive narrative patterns that reduced contextual variety. Based on the combined quantitative and qualitative assessment, we selected Qwen3-32B, which achieved the best balance of \textit{Narrative Quality}, \textit{Discourse Coherence}, and \textit{Coreference Density}. Figures detailing all results are in Appendix~\ref{app:eval_model_selection}.


\textbf{Prompt Design.} Using Qwen3-32B, we systematically varied two prompt dimensions, evaluating each on the development set using the same metrics described above. (1) \textit{Number of characters}: We tested prompts specifying 2, 3, or any number of entities. Unconstrained prompts yielded the most coherent narratives, as fixed character counts introduced unnecessary complexity. (2) 
\textit{Narrative length}: We tested lengths of 5, 10, 15 and 20 sentences. Narratives of 10 sentences offered the best trade-off between coherence and coreference density, with longer narratives suffering from grammatical and consistency degradation. The final prompt combining all optimal settings are provided in Appendix~\ref{app:final_narrative}. Detailed results for all design dimensions are included in Appendix~\ref{app:eval_prompt_design}.

\textbf{Final Dataset Generation.} Using the final prompt and Qwen3-32B, we generated narratives for all 5,994 instances in the UFET crowd-annotated set. Each narrative is a 10-sentence passage that embeds the original sentence verbatim, with no constraint on the number of characters. We generate two variants: Narrative-UFET-\textit{Change}, in which the prompt instructs the generator to shift the type of the target entity across the narrative, and Narrative-UFET-\textit{Maintain}, in which the prompt instructs it to hold the type constant. Crucially, the gold types from UFET are never shown to the generator. The prompt specifies only whether the type should change or remain stable, not which type to use. This ensures that any improvements observed under Narrative-UFET reflect richer discourse context rather than direct exposure to the labels.

\subsection{Human Validation}

To verify narrative quality beyond automated metrics, we conducted a human evaluation on 100 randomly sampled narratives from the test set on both Narrative-UFET-\textit{Change} and Narrative-UFET-\textit{Maintain}. There were four different annotators in total, two annotators for each dataset. Annotators rated each narrative on a 5-point Likert scale across six dimensions: grammar, creativity, consistency, plot, context-to-story coherence, and story-internal coherence. Rubric definitions are provided in Appendix~\ref{app:rubric}. A pilot study on five examples confirmed calibration, with annotators agreeing or differing by at most 1 point on all dimensions.

We report inter-annotator agreement using Gwet's AC2~\cite{gwet2014handbook}. This agreement method was utilized due to the heavy right skew from all annotators scores (every annotator gave majority scores of 3 or higher in all dimensions). Agreement was substantial to almost perfect on five of six dimensions: grammar (AC2 = 0.80, 0.84), context-to-story coherence (AC2 = 0.86, 0.84), story-internal coherence (AC2 = 0.87, 0.81), consistency (AC2 = 0.89, 0.78), and plot (AC2 = 0.86, 0.71). Creativity showed substantial agreement on Narrative-UFET-\textit{Change} (AC2 = 0.7365) and moderate agreement on Narrative-UFET-\textit{Maintain} (AC2 = 0.45), reflecting its inherently subjective nature. Average scores across both annotators ranged from 3.56 to 4.80, indicating that the generated narratives are of consistently high quality. Per-dimension scores and annotator breakdowns in Appendix~\ref{app:annotator_results}.

%% file: latex/4_experiments.tex
\section{Experiments}

We evaluate whether Narrative-UFET improves entity typing performance with respect to sentence-level UFET, particularly for long-tail entities. Following \citet{deshmukh-etal-2025-entities}, we partition the UFET test set into four bins. Details on the bin split and internet search hits are shown in Appendix~\ref{app:bin_split_details}. We evaluate using both masked language model (MLM) and causal language model (CLM) approaches. All scoring is done following the UFET guidelines~\cite{choi-etal-2018-ultra}.



\begin{figure}
    \centering
    \includegraphics[width=1.0\linewidth]{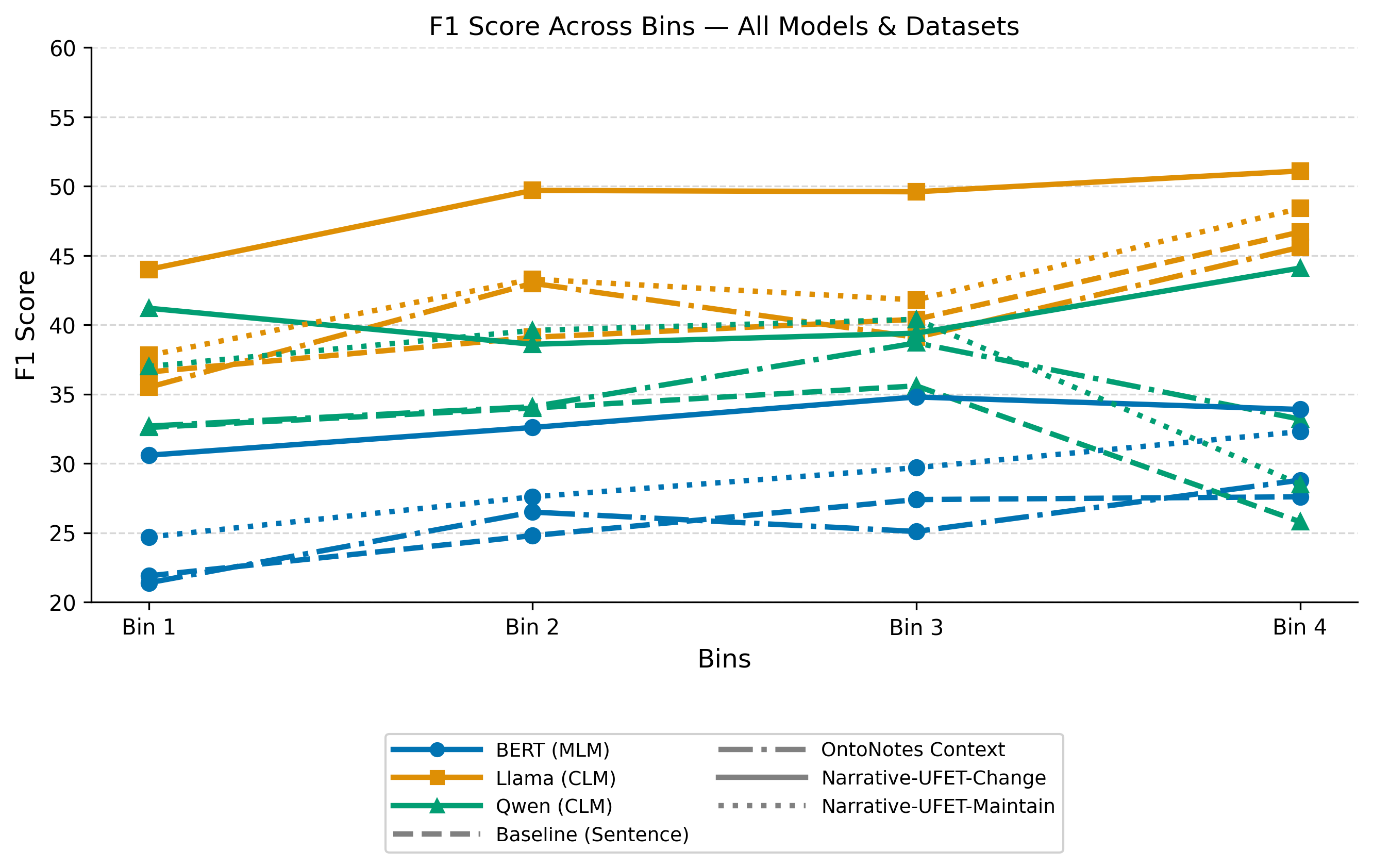}
    \caption{Type F1 Scores Across UFET Bins: MLM and CLM performance between Narrative-UFET, sentence-level, and OntoNotes 5.0 context.}
    \label{fig:f1_score_mlm_vs_causal}
\end{figure}

\textbf{Narrative context improves typing performance.} We adopt the MLM and CLM setups of \citet{deshmukh-etal-2025-entities}; details are in Appendices~\ref{app:entity_typing_hearst} and~\ref{app:entity_typing_prompt_nar}. For CLMs, we run each condition five times and report the mean to account for model variability. Figure~\ref{fig:f1_score_mlm_vs_causal} compares performance between the sentence-level baseline and Narrative-UFET across both setups. Narrative-UFET yields consistent improvements across all bins, with CLMs reaching higher overall performance than MLMs, and Qwen showing a larger relative gain. Across both model types, Narrative-UFET-\textit{Change} substantially outperforms Narrative-UFET-\textit{Maintain}, suggesting that changing the type across the narrative produces a richer typing signal.

\textbf{Synthetic narratives outperform real context.} To check that our synthetic narratives capture properties of real discourse, we run both MLM and CLM evaluations using the original OntoNotes 5.0 contexts \cite{DBLP:journals/corr/GillickLGKH14}. Figure~\ref{fig:f1_score_mlm_vs_causal} shows that Narrative-UFET-\textit{Change} outperforms OntoNotes context across both model types, while OntoNotes performance sits between Standard-UFET and Narrative-UFET-\textit{Maintain}. This indicates that the synthetic narratives preserve properties of naturally occurring context while the controlled type-shift design in \textit{Change} surfaces typing signal that real text leaves implicit.


Results for additional model families and sizes, and additional analyses in Appendix~\ref{app:entity_typing_results}.

%% file: latex/5_conclusion.tex
\section{Conclusion and Future Work}

We presented \textbf{Narrative-UFET}, a controlled narrative-level extension of UFET in which each entity mention is paired with an automatically generated narrative built around it. Our hypothesis was that synthesizing narratives lets us isolate the effect of specific discourse properties. To study this, we constructed two variants that differ in whether the entity's type is held constant (\textit{Maintain}) or shifts (\textit{Change}) across the narrative. We found that narrative context yields consistent gains on long-tail entities over sentence-level baselines, with the \textit{Change} variant providing the stronger typing signal, and outperforming naturally occurring contexts, suggesting that controlled construction can surface signal that real text leaves implicit.

The type-consistency contrast studied here is one of many discourse properties that controlled narrative construction makes accessible. Coreference density, interactions between co-occurring entities, the granularity of type shifts, and the distribution of typing evidence across sentences are all dimensions that Narrative-UFET could be extended along, and each may explain a different part of the gap between current models and the typing signal available in discourse. We see this as the main direction opened by our work: not just longer context for entity typing, but a systematic investigation of which discourse properties carry typing signal and how models can exploit them.

%% file: latex/6_limitations.tex
\section*{Limitations}

We identify six main limitations. (1) Our case study varies only one property of the narrative: whether the entity's type is held constant or shifts. Other discourse properties (e.g., coreference density, inter-entity interactions, the distribution of typing evidence across sentences) are likely to carry typing signal as well, and we leave their systematic study to future work. (2) While we validate narrative quality through automated and human evaluation, and compare against naturally occurring contexts from OntoNotes, synthetic narratives may deviate from natural discourse distributions in ways that our evaluation does not capture. Extending Narrative-UFET with retrieved or human-authored narratives is an important direction. (3) We attribute the stronger performance of the \textit{Change} variant to richer typing signal from type variation across discourse, but we do not control for confounding factors such as token count, lexical diversity, or the number of distinct types mentioned. The relative contribution of these factors remains open. (4) Our experiments cover one MLM (BERT) and two CLMs (Llama3.3-70B, Qwen3-32B) on a single base typing dataset (UFET, English, crowd-annotated). Findings may not transfer to other typing schemes, languages, or model families.
(5) For computational reasons, all models in Section \ref{sec:3} were run in quantized form. Quantization can affect both narrative generation quality and typing predictions. We did not systematically measure these effects. (6) Our qualitative review during model selection was conducted by a single annotator without formal guidelines. While the final dataset is also validated through inter-annotator human evaluation with formal rubrics, the model-selection stage relied on lighter-weight judgment.

%% file: latex/7_ethical-considerations.tex
\section*{Ethical Considerations}

To the best of our knowledge, this work does not raise any ethical concerns. All information required to replicate our experiments is provided in the paper and the appendices. We use only open-source language models with publicly available and versioned weights. Importantly, we ensure separation between the models used for narrative generation, narrative evaluation, and entity typing evaluation, avoiding circular dependencies that could inflate results. Additional plots and implementation details are provided in the appendix.

%% file: latex/0_acl_latex.bbl
\begin{thebibliography}{28}
\providecommand{\natexlab}[1]{#1}

\bibitem[{Chhun et~al.(2022)Chhun, Colombo, Suchanek, and Clavel}]{chhun-etal-2022-human}
Cyril Chhun, Pierre Colombo, Fabian~M. Suchanek, and Chlo{\'e} Clavel. 2022.
\newblock \href {https://aclanthology.org/2022.coling-1.509/} {Of human criteria and automatic metrics: A benchmark of the evaluation of story generation}.
\newblock In \emph{Proceedings of the 29th International Conference on Computational Linguistics}, pages 5794--5836, Gyeongju, Republic of Korea. International Committee on Computational Linguistics.

\bibitem[{Choi et~al.(2018)Choi, Levy, Choi, and Zettlemoyer}]{choi-etal-2018-ultra}
Eunsol Choi, Omer Levy, Yejin Choi, and Luke Zettlemoyer. 2018.
\newblock \href {https://doi.org/10.18653/v1/P18-1009} {Ultra-fine entity typing}.
\newblock In \emph{Proceedings of the 56th Annual Meeting of the Association for Computational Linguistics (Volume 1: Long Papers)}, pages 87--96, Melbourne, Australia. Association for Computational Linguistics.

\bibitem[{Dai et~al.(2021)Dai, Song, and Wang}]{dai-etal-2021-ultra}
Hongliang Dai, Yangqiu Song, and Haixun Wang. 2021.
\newblock \href {https://doi.org/10.18653/v1/2021.acl-long.141} {Ultra-fine entity typing with weak supervision from a masked language model}.
\newblock In \emph{Proceedings of the 59th Annual Meeting of the Association for Computational Linguistics and the 11th International Joint Conference on Natural Language Processing (Volume 1: Long Papers)}, pages 1790--1799, Online. Association for Computational Linguistics.

\bibitem[{Deshmukh et~al.(2025)Deshmukh, Umadi, Srinivas, and Pacheco}]{deshmukh-etal-2025-entities}
Advait Deshmukh, Ashwin Umadi, Dananjay Srinivas, and Maria~Leonor Pacheco. 2025.
\newblock \href {https://doi.org/10.18653/v1/2025.starsem-1.15} {All entities are not created equal: Examining the long tail for ultra-fine entity typing}.
\newblock In \emph{Proceedings of the 14th Joint Conference on Lexical and Computational Semantics (*SEM 2025)}, pages 189--201, Suzhou, China. Association for Computational Linguistics.

\bibitem[{Ding et~al.(2021)Ding, Xu, Chen, Wang, Han, Xie, Zheng, and Liu}]{ding-etal-2021-nerd}
Ning Ding, Guangwei Xu, Yulin Chen, Xiaobin Wang, Xu~Han, Pengjun Xie, Haitao Zheng, and Zhiyuan Liu. 2021.
\newblock \href {https://doi.org/10.18653/v1/2021.acl-long.248} {Few-{NERD}: A few-shot named entity recognition dataset}.
\newblock In \emph{Proceedings of the 59th Annual Meeting of the Association for Computational Linguistics and the 11th International Joint Conference on Natural Language Processing (Volume 1: Long Papers)}, pages 3198--3213, Online. Association for Computational Linguistics.

\bibitem[{Eldan and Li(2023)}]{eldan2023tinystoriessmalllanguagemodels}
Ronen Eldan and Yuanzhi Li. 2023.
\newblock \href {https://arxiv.org/abs/2305.07759} {Tinystories: How small can language models be and still speak coherent english?}
\newblock \emph{Preprint}, arXiv:2305.07759.

\bibitem[{Gillick et~al.(2014{\natexlab{a}})Gillick, Lazic, Ganchev, Kirchner, and Huynh}]{DBLP:journals/corr/GillickLGKH14}
Dan Gillick, Nevena Lazic, Kuzman Ganchev, Jesse Kirchner, and David Huynh. 2014{\natexlab{a}}.
\newblock \href {https://arxiv.org/abs/1412.1820} {Context-dependent fine-grained entity type tagging}.
\newblock \emph{CoRR}, abs/1412.1820.

\bibitem[{Gillick et~al.(2014{\natexlab{b}})Gillick, Lazic, Ganchev, Kirchner, and Huynh}]{Gillick2014ContextDependentFE}
Daniel Gillick, Nevena Lazic, Kuzman Ganchev, Jesse Kirchner, and David Huynh. 2014{\natexlab{b}}.
\newblock \href {https://api.semanticscholar.org/CorpusID:9836000} {Context-dependent fine-grained entity type tagging}.
\newblock \emph{ArXiv}, abs/1412.1820.

\bibitem[{Goldfarb-Tarrant et~al.(2020)Goldfarb-Tarrant, Chakrabarty, Weischedel, and Peng}]{goldfarb-tarrant-etal-2020-content}
Seraphina Goldfarb-Tarrant, Tuhin Chakrabarty, Ralph Weischedel, and Nanyun Peng. 2020.
\newblock \href {https://doi.org/10.18653/v1/2020.emnlp-main.351} {Content planning for neural story generation with aristotelian rescoring}.
\newblock In \emph{Proceedings of the 2020 Conference on Empirical Methods in Natural Language Processing (EMNLP)}, pages 4319--4338, Online. Association for Computational Linguistics.

\bibitem[{Gwet(2014)}]{gwet2014handbook}
K.L. Gwet. 2014.
\newblock \href {https://books.google.com/books?id=fac9BQAAQBAJ} {\emph{Handbook of Inter-Rater Reliability, 4th Edition: The Definitive Guide to Measuring The Extent of Agreement Among Raters}}.
\newblock Advanced Analytics, LLC.

\bibitem[{Harel-Canada et~al.(2024)Harel-Canada, Zhou, Muppalla, Yildiz, Kim, Sahai, and Peng}]{harel-canada-etal-2024-measuring}
Fabrice~Y Harel-Canada, Hanyu Zhou, Sreya Muppalla, Zeynep~Senahan Yildiz, Miryung Kim, Amit Sahai, and Nanyun Peng. 2024.
\newblock \href {https://doi.org/10.18653/v1/2024.emnlp-main.953} {Measuring psychological depth in language models}.
\newblock In \emph{Proceedings of the 2024 Conference on Empirical Methods in Natural Language Processing}, pages 17162--17196, Miami, Florida, USA. Association for Computational Linguistics.

\bibitem[{Koch et~al.(2014)Koch, Gilmer, Soderland, and Weld}]{koch-etal-2014-type}
Mitchell Koch, John Gilmer, Stephen Soderland, and Daniel~S. Weld. 2014.
\newblock \href {https://doi.org/10.3115/v1/D14-1203} {Type-aware distantly supervised relation extraction with linked arguments}.
\newblock In \emph{Proceedings of the 2014 Conference on Empirical Methods in Natural Language Processing ({EMNLP})}, pages 1891--1901, Doha, Qatar. Association for Computational Linguistics.

\bibitem[{Li et~al.(2024{\natexlab{a}})Li, Li, Sun, Wang, Zhang, Wang, and Pan}]{li-etal-2024-llms}
Jinyuan Li, Han Li, Di~Sun, Jiahao Wang, Wenkun Zhang, Zan Wang, and Gang Pan. 2024{\natexlab{a}}.
\newblock \href {https://doi.org/10.18653/v1/2024.findings-acl.76} {{LLM}s as bridges: Reformulating grounded multimodal named entity recognition}.
\newblock In \emph{Findings of the Association for Computational Linguistics: ACL 2024}, pages 1302--1318, Bangkok, Thailand. Association for Computational Linguistics.

\bibitem[{Li et~al.(2024{\natexlab{b}})Li, Hu, King, and Leung}]{li-etal-2024-integration}
Muzhi Li, Minda Hu, Irwin King, and Ho-fung Leung. 2024{\natexlab{b}}.
\newblock \href {https://doi.org/10.18653/v1/2024.naacl-long.369} {The integration of semantic and structural knowledge in knowledge graph entity typing}.
\newblock In \emph{Proceedings of the 2024 Conference of the North American Chapter of the Association for Computational Linguistics: Human Language Technologies (Volume 1: Long Papers)}, pages 6625--6638, Mexico City, Mexico. Association for Computational Linguistics.

\bibitem[{Li et~al.(2023)Li, Bouraoui, and Schockaert}]{li-etal-2023-ultra}
Na~Li, Zied Bouraoui, and Steven Schockaert. 2023.
\newblock \href {https://doi.org/10.18653/v1/2023.findings-emnlp.786} {Ultra-fine entity typing with prior knowledge about labels: A simple clustering based strategy}.
\newblock In \emph{Findings of the Association for Computational Linguistics: EMNLP 2023}, pages 11744--11756, Singapore. Association for Computational Linguistics.

\bibitem[{Ling et~al.(2015)Ling, Singh, and Weld}]{ling-etal-2015-design}
Xiao Ling, Sameer Singh, and Daniel~S. Weld. 2015.
\newblock \href {https://doi.org/10.1162/tacl_a_00141} {Design challenges for entity linking}.
\newblock \emph{Transactions of the Association for Computational Linguistics}, 3:315--328.

\bibitem[{Ling and Weld(2021)}]{Ling_Weld_2021}
Xiao Ling and Daniel Weld. 2021.
\newblock \href {https://doi.org/10.1609/aaai.v26i1.8122} {Fine-grained entity recognition}.
\newblock \emph{Proceedings of the AAAI Conference on Artificial Intelligence}, 26(1):94--100.

\bibitem[{Naismith et~al.(2023)Naismith, Mulcaire, and Burstein}]{naismith-etal-2023-automated}
Ben Naismith, Phoebe Mulcaire, and Jill Burstein. 2023.
\newblock \href {https://doi.org/10.18653/v1/2023.bea-1.32} {Automated evaluation of written discourse coherence using {GPT}-4}.
\newblock In \emph{Proceedings of the 18th Workshop on Innovative Use of NLP for Building Educational Applications (BEA 2023)}, pages 394--403, Toronto, Canada. Association for Computational Linguistics.

\bibitem[{Onoe and Durrett(2020)}]{onoe-durrett-2020-interpretable}
Yasumasa Onoe and Greg Durrett. 2020.
\newblock \href {https://doi.org/10.18653/v1/2020.findings-emnlp.54} {Interpretable entity representations through large-scale typing}.
\newblock In \emph{Findings of the Association for Computational Linguistics: EMNLP 2020}, pages 612--624, Online. Association for Computational Linguistics.

\bibitem[{Pan et~al.(2022)Pan, Wei, and Zhu}]{ijcai2022p599}
Weiran Pan, Wei Wei, and Feida Zhu. 2022.
\newblock \href {https://doi.org/10.24963/ijcai.2022/599} {Automatic noisy label correction for fine-grained entity typing}.
\newblock In \emph{Proceedings of the Thirty-First International Joint Conference on Artificial Intelligence, {IJCAI-22}}, pages 4317--4323. International Joint Conferences on Artificial Intelligence Organization.
\newblock Main Track.

\bibitem[{Raffel et~al.(2020)Raffel, Shazeer, Roberts, Lee, Narang, Matena, Zhou, Li, and Liu}]{JMLR:v21:20-074}
Colin Raffel, Noam Shazeer, Adam Roberts, Katherine Lee, Sharan Narang, Michael Matena, Yanqi Zhou, Wei Li, and Peter~J. Liu. 2020.
\newblock \href {http://jmlr.org/papers/v21/20-074.html} {Exploring the limits of transfer learning with a unified text-to-text transformer}.
\newblock \emph{Journal of Machine Learning Research}, 21(140):1--67.

\bibitem[{Riedel et~al.(2010)Riedel, Yao, and McCallum}]{Riedel2010ModelingRA}
Sebastian Riedel, Limin Yao, and Andrew McCallum. 2010.
\newblock \href {https://api.semanticscholar.org/CorpusID:2386383} {Modeling relations and their mentions without labeled text}.
\newblock In \emph{ECML/PKDD}.

\bibitem[{Tang et~al.(2024)Tang, Chen, Lin, and Li}]{TANG2024e34262}
Xiaoyi Tang, Hongwei Chen, Daoyu Lin, and Kexin Li. 2024.
\newblock \href {https://doi.org/10.1016/j.heliyon.2024.e34262} {Harnessing llms for multi-dimensional writing assessment: Reliability and alignment with human judgments}.
\newblock \emph{Heliyon}, 10(14):e34262.

\bibitem[{Tian et~al.(2024)Tian, Huang, Liu, Jiang, Spangher, Chen, May, and Peng}]{tian-etal-2024-large-language}
Yufei Tian, Tenghao Huang, Miri Liu, Derek Jiang, Alexander Spangher, Muhao Chen, Jonathan May, and Nanyun Peng. 2024.
\newblock \href {https://doi.org/10.18653/v1/2024.emnlp-main.978} {Are large language models capable of generating human-level narratives?}
\newblock In \emph{Proceedings of the 2024 Conference on Empirical Methods in Natural Language Processing}, pages 17659--17681, Miami, Florida, USA. Association for Computational Linguistics.

\bibitem[{Wang et~al.(2024)Wang, Zhu, Zheng, Li, Xu, He, Liu, Yu, and Chen}]{wang-etal-2024-granular}
Ziqi Wang, Chen Zhu, Zhi Zheng, Xinhang Li, Tong Xu, Yongyi He, Qi~Liu, Ying Yu, and Enhong Chen. 2024.
\newblock \href {https://doi.org/10.18653/v1/2024.findings-emnlp.183} {Granular entity mapper: Advancing fine-grained multimodal named entity recognition and grounding}.
\newblock In \emph{Findings of the Association for Computational Linguistics: EMNLP 2024}, pages 3211--3226, Miami, Florida, USA. Association for Computational Linguistics.

\bibitem[{Weber et~al.(2024)Weber, Fu, Anthony, Oren, Adams, Alexandrov, Lyu, Nguyen, Yao, Adams, Athiwaratkun, Chalamala, Chen, Ryabinin, Dao, Liang, Ré, Rish, and Zhang}]{weber2024redpajamaopendatasettraining}
Maurice Weber, Daniel Fu, Quentin Anthony, Yonatan Oren, Shane Adams, Anton Alexandrov, Xiaozhong Lyu, Huu Nguyen, Xiaozhe Yao, Virginia Adams, Ben Athiwaratkun, Rahul Chalamala, Kezhen Chen, Max Ryabinin, Tri Dao, Percy Liang, Christopher Ré, Irina Rish, and Ce~Zhang. 2024.
\newblock \href {https://arxiv.org/abs/2411.12372} {Redpajama: an open dataset for training large language models}.
\newblock \emph{Preprint}, arXiv:2411.12372.

\bibitem[{Yang et~al.(2022)Yang, Tian, Peng, and Klein}]{yang-etal-2022-re3}
Kevin Yang, Yuandong Tian, Nanyun Peng, and Dan Klein. 2022.
\newblock \href {https://doi.org/10.18653/v1/2022.emnlp-main.296} {Re3: Generating longer stories with recursive reprompting and revision}.
\newblock In \emph{Proceedings of the 2022 Conference on Empirical Methods in Natural Language Processing}, pages 4393--4479, Abu Dhabi, United Arab Emirates. Association for Computational Linguistics.

\bibitem[{Zhu et~al.(2015)Zhu, Kiros, Zemel, Salakhutdinov, Urtasun, Torralba, and Fidler}]{DBLP:journals/corr/ZhuKZSUTF15}
Yukun Zhu, Ryan Kiros, Richard~S. Zemel, Ruslan Salakhutdinov, Raquel Urtasun, Antonio Torralba, and Sanja Fidler. 2015.
\newblock \href {https://arxiv.org/abs/1506.06724} {Aligning books and movies: Towards story-like visual explanations by watching movies and reading books}.
\newblock \emph{CoRR}, abs/1506.06724.

\end{thebibliography}
